\documentclass[runningheads]{llncs}

\usepackage[mobile]{eccv}

\usepackage{eccvabbrv}

\usepackage{multirow}
\usepackage[table,xcdraw]{xcolor}
\usepackage{array}
\usepackage{makecell}
\usepackage{caption}
\usepackage{graphicx}
\usepackage{pifont}
\usepackage{booktabs}
\usepackage{calc}
\usepackage{textcomp}

\usepackage[accsupp]{axessibility}  

\usepackage[pagebackref,breaklinks,colorlinks,citecolor=eccvblue]{hyperref}
\usepackage{hyperref}

\usepackage{orcidlink}

\newcommand{\benchmark}{M\textsuperscript{3}-Verse}

\newcommand{\simsubset}{M\textsuperscript{3}-Verse-Sim}
\newcommand{\realsubset}{M\textsuperscript{3}-Verse-Real}

\begin{document}

\title{\textbf{\benchmark}: A ``Spot the Difference'' Challenge for Large Multimodal Models} 


\author{
  Kewei Wei\inst{1}$^\dagger$ \and
  Bocheng Hu\inst{1}$^\dagger$ \and
  Jie Cao\inst{1} \and
  Xiaohan Chen\inst{1} \and
  Zhengxi Lu\inst{1} \and
  Wubing Xia\inst{1} \and
  Weili Xu\inst{1} \and
  Jiaao Wu\inst{1} \and
  Junchen He\inst{1} \and
  Mingyu Jia\inst{1} \and
  Ciyun Zhao\inst{1} \and
  Ye Sun\inst{1} \and
  Yizhi Li\inst{1} \and
  Zhonghan Zhao\inst{1,2} \and
  Jian Zhang\inst{3} \and
  Gaoang Wang\inst{1,2}$^\ast$
}

\authorrunning{K.~Wei et al.}

\institute{Zhejiang University, China \and
Shanghai AI Lab, China \and
Hangzhou Normal University, China}

\footnotetext{$^\dagger$ These authors contributed equally to this work.}
\footnotetext{$^\ast$ Corresponding author. Email: \texttt{gaoangwang@intl.zju.edu.cn}}
\footnotetext{Important authors' emails: 
  Kewei Wei: \texttt{12421275@zju.edu.cn}, 
  Bocheng Hu: \texttt{bocheng.25@intl.zju.edu.cn}, 
  Jie Cao: \texttt{jie.25@intl.zju.edu.cn}, 
  Xiaohan Chen: \texttt{xiaohan.25@intl.zju.edu.cn}, 
  Zhengxi Lu: \texttt{zhengxilu@zju.edu.cn}, 
  Weili Xu: \texttt{weili.23@intl.zju.edu.cn}, 
  Zhonghan Zhao: \texttt{zhaozhonghan@zju.edu.cn}, 
}

\maketitle

\begin{abstract}
    Modern Large Multimodal Models (LMMs) have demonstrated extraordinary ability in static image and single-state spatial-temporal understanding. However, their capacity to comprehend the dynamic changes of objects within a shared spatial context between two distinct video observations remains largely unexplored. In this paper, we introduce \textbf{\benchmark}, a \textbf{M}ulti-Modal, \textbf{M}ulti-State, and \textbf{M}ulti-Dimensional benchmark evaluating this capability through paired videos of scenes before and after state changes. The benchmark features two rigorous subsets spanning over 50 subtasks. \textbf{\simsubset} contains \textit{270 simulated scenes} with \textit{2,932 filtered QAs}, and \textbf{\realsubset} provides \textit{29 physical scenes} with \textit{552 human annotated QAs}. Evaluating \textbf{16 state-of-the-art LMMs} reveals severe limitations in tracking state transitions. To diagnose these failures, we design a \textbf{Text-Oracle Diagnostic Probe} by explicitly serializing visual streams into text. We demonstrate that vision encoders successfully capture state changes, yet models fail to retain these sparse visual cues across long temporal contexts. \textbf{\benchmark} thus provides a challenging new testbed to catalyze the development of next-generation models with a more holistic understanding of our dynamic visual world.
  \keywords{State Tracking Reasoning \and Video Understanding \and Multimodal Benchmark}
\end{abstract}
\section{Introduction}

\begin{figure*}[h]
  \centering
    \includegraphics[width=\linewidth]{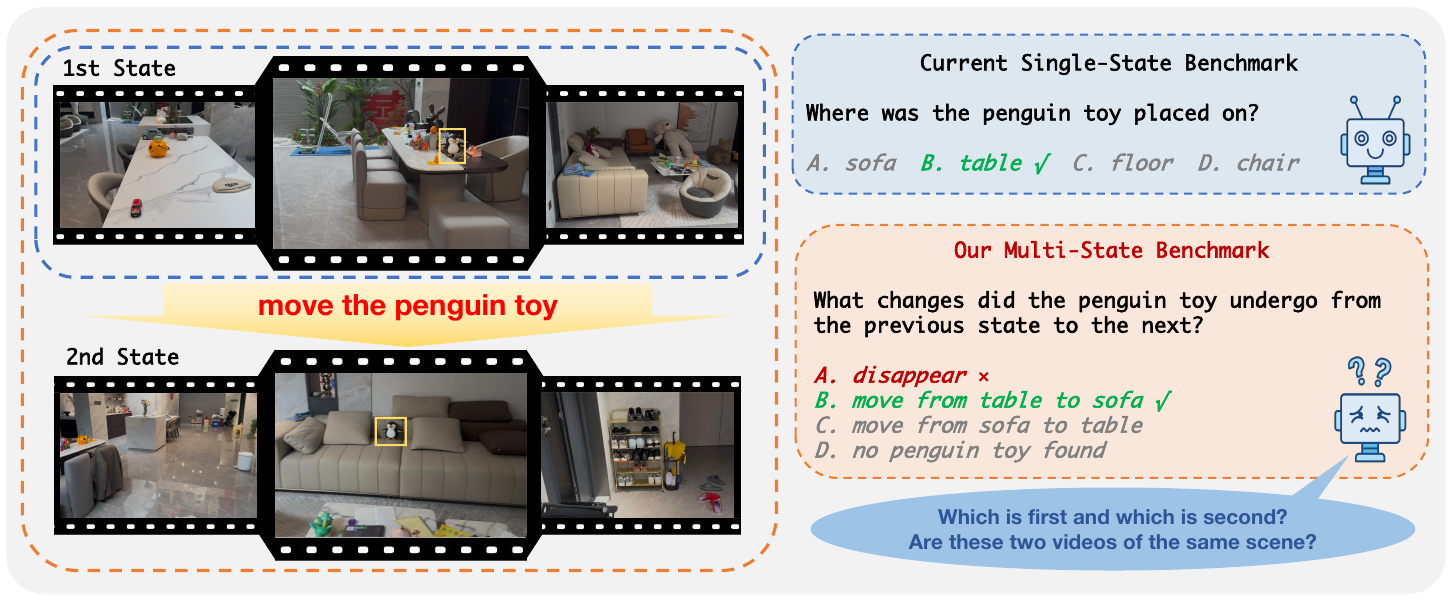}
    \caption{Limitation of current LMM benchmarks. Existing benchmarks concentrate on evaluating a model's performance on an isolated state, neglecting to test its ability to understand changes across different states. Furthermore, the lack of relevant data and benchmarks has impeded the development of model capabilities in this area. This work is designed to fill that gap.}
    \label{fig:teaser}
\end{figure*}

Biological intelligence naturally demonstrates a vital capacity to perceive environmental changes by contrasting past and present scenes to detect subtle differences\cite{changing_experience}. For instance, animals spot disturbed vegetation to evade predators or track prey, and humans refine sequential actions to learn new skills. A truly advanced cognitive agent should continuously improve itself by perceiving and comparing these changes. This capability forms the foundation of higher-order cognition. Without this fundamental capacity for change detection, no meaningful learning can occur. While recent state-of-the-art large multimodal models (LMMs)~\cite{qwen3-vl,InternVL3.5,MiniCPM-V-4.5,Video-XL-2,VideoLLaMA3,Llava-OneVision-1.5} have demonstrated remarkable advancements, their focus has predominantly been on interpreting static images or online tracking within continuous video streams. Existing spatial-temporal benchmarks \cite{VSI-Bench,Sti-bench,viewspatial,spatial-mm,spatialrgpt} generally assess observations in single continuous videos but neglect cross-state comparative reasoning. Comparing disconnected videos imposes a substantially higher cognitive load than traditional visual question answering. An intelligent agent must carefully observe the initial video, establish a working memory of the spatial layout and object states, and then cross-reference the subsequent video to locate changes.

To address this critical gap, we introduce \textbf{\benchmark}, a novel benchmark designed to evaluate the understanding of discrete state changes. Its two-state structure transcends simple recognition, requiring models to perform a fine-grained semantic comparison and infer the transformations that connect the two scenes. We offer the following principal contributions:
\begin{itemize}
    \item \textbf{A Novel High-Quality Diagnostic Benchmark:} We introduce \textbf{\benchmark}, the first fine-grained benchmark utilizing a challenging two-state paired-video paradigm to evaluate comparative reasoning about state transitions. It features two shortcut-resistant subsets covering over 50 subtasks and 4 core capabilities. \textbf{\simsubset} provides comprehensive diagnostic profiling via 270 simulated scenes and 2,932 rigorously filtered questions, while \textbf{\realsubset} assesses real-world robustness with 29 physical scenes and 552 strictly human-annotated questions.
﻿
    \item \textbf{In-depth Model Analysis:} Our in-depth evaluation of \textbf{16 leading LMMs} and \textbf{6 text-only LLMs} reveals critical weakness in comparative spatial-temporal reasoning. The results demonstrate a stark performance gap between current SOTA models and human cognition, highlighting that the fundamental capacity to track and interpret discrete state transitions across distinct observations remains a challenge for today's LMMs.
﻿
    \item \textbf{Text-Oracle Diagnostic Analysis:} We decouple visual perception from logical reasoning via a novel Text-Oracle probe. This analysis proves that while vision encoders successfully capture state changes, these sparse cues are diluted over long temporal contexts. Furthermore, we reveal that the ultimate upper bound for this task is heavily dictated by the logical reasoning ceiling of the LLM backbone.
\end{itemize}

By open-sourcing the \textbf{\benchmark}, we provide the community with a vital tool to advance the next generation of AI systems capable of understanding and reasoning about our dynamic world.
\section{Related Works}
\subsection{Large Multimodal Models}
The development of LMMs has markedly advanced AI's ability to perceive the world. Early works~\cite{clip,blip,flamingo} mapped visual features to text embeddings, leading to models~\cite{LLaVA,qwenvl} that bridged visual encoders with LLMs for impressive visual question answering. This architecture was then extended from static images to short video understanding~\cite{videollama,videochat,videollava,llavanext,videochatgpt}. To push the boundaries of the model's capabilities, recent research has focused on long-term video understanding, tackling the challenge of long contexts with methods like token merging strategies~\cite{moviechat,longvlm} and innovative architectures~\cite{InternVideo2.5,auroralong,MiniCPM-V-4.5,Video-XL-2}. Beyond extending context length, a new frontier aims to enhance the reasoning capabilities of LMMs via post-training techniques like reinforcement learning (RL)~\cite{Video-R1,WeThink, glm4.1vthinking}.

While these models have made significant progress in video processing, their ability to understand fine-grained changes in discrete multi-state videos remains unexplored. We introduce a benchmark, concentrating on this overlooked dimension, multi-state video understanding, and utilize it to evaluate how well current SOTA models can identify the state transitions of key objects or scenes at different points in time.

\subsection{Benchmarks for LMMs}
To evaluate the visual understanding capabilities of LMMs, a large number of benchmarks have been proposed. Initial efforts aggregated traditional datasets for foundational image understanding tasks~\cite{mmbench} and general video understanding~\cite{mmbench-video}. More recently, the focus has shifted to fine-graind perception capabilities. Specialized benchmarks now assess fine-grained spatial relationships~\cite{spatial-mm,VSI-Bench,viewspatial}, temporal and state-transition logic~\cite{visualtrans,vstar,Sti-bench,svbench}. Beyond assessing specific abilities, the landscape of video understanding benchmarks has diversified. Some benchmarks~\cite{OST-Bench,svbench} are designed to test a model's ability to acquire information incrementally online and reason with it. Some~\cite{H2VU-Benchmark,OST-Bench,mtvideo-bench} focus on hierarchical analysis of long-form videos, and others expand into cross-video contexts~\cite{CVBench,mtvideo-bench,svbench}. 

Research like SAT~\cite{SAT} has begun to investigate state transformations, but its scope is restricted to comparing static images. Overall, this direction remains in a preliminary stage of exploration. Our work is designed to fill this void by providing a targeted assessment of a model's ability to perceive how objects or scenes change between distinct states, offering a crucial supplement to the current evaluation landscape.

\newcommand{\cmark}{{\color{green!70!black}\ding{51}}} 
\newcommand{\xmark}{{\color{red}\ding{55}}} 

\begin{table*}[t]
\centering
\small
\renewcommand{\arraystretch}{1.15}
\caption{ \textbf{Comparison with existing visual QA datasets and benchmarks.} 
I = Image, V = Video, Q = Question, M = Multi, S = Simulated data, R = Real data. Dashes(--) indicate unreported values. \underline{Underlined} indicate not publicly available.}
\resizebox{\textwidth}{!}{%
\begin{tabular}{l|c|c|c|c|c|c|c|c}
\toprule
\textbf{Source} & \textbf{\# Scenes} & \textbf{\# Scale} & \textbf{\# Tasks} & \textbf{\# Qs} & \textbf{M-V/Q} & \textbf{M-State} & \textbf{Extra Ann.} & \textbf{Source} \\ 
\midrule
\multicolumn{9}{l}{\textit{Datasets}} \\
\midrule
\underline{VST-P}~\cite{VST} & -- & 4.1M+ Is & 19 & 4.1M+ & \xmark & \xmark & -- & -- \\
\underline{VST-R}~\cite{VST} & -- & 135,000 Is & 14 & 135,000 & \xmark & \xmark & -- & -- \\
VSI-590K~\cite{Cambrian-S} & -- & 44,858 Is + 5,963 Vs & 5 & 590,667 & \xmark & \xmark & \xmark & S + R \\
SAT~\cite{SAT} & -- & 8K Is & 6 & 306K & \xmark & \cmark & \xmark & S \\
SIMS-V~\cite{SIMS-V} & 1,261 & 2,507 Vs & 9 & 203,048 & \xmark & \xmark & \cmark & S \\
\midrule
\multicolumn{9}{l}{\textit{Benchmarks}} \\
\midrule
Spatial-MM~\cite{spatial-mm} & -- & 322 Is & 36 & 2,622 & \xmark & \xmark & \xmark & R \\
SpatialRGPT~\cite{spatialrgpt} & -- & 1,406 Is & 12 & 1,406 & \xmark & \xmark & \cmark & S + R \\
ViewSpatial-Bench~\cite{viewspatial} & 279 & 279 Vs (6,150 Is) & 5 & 5,712 & \xmark & \xmark & \cmark & S + R \\
VisualTrans~\cite{visualtrans} & 12 & 1,193 Is & 6 & 497 & \xmark & \cmark & \xmark & S \\
SVBench~\cite{svbench} & -- & 1,353 Vs & 9 & 49,979 & \xmark & \cmark & \xmark & R \\
H\textsuperscript{2}VU-Benchmark~\cite{H2VU-Benchmark} & -- & 5,902 Vs & 47 & 10,183 & \xmark & \cmark & \xmark & R \\
MT-Video-Bench~\cite{mtvideo-bench} & -- & 135 Vs & 23 & 5,805 & \xmark & \cmark & \xmark & R \\
STI-Bench~\cite{Sti-bench} & -- & 300 Vs & 8 & 2,000+ & \xmark & \cmark & \xmark & S + R \\
OST-Bench~\cite{OST-Bench} & 1,386 & 1,386 Vs & 15 & 10,165 & \xmark & \xmark & \xmark & S \\
VSI-bench~\cite{VSI-Bench} & 288 & 288 Vs & 8 & 5,000 & \xmark & \xmark & \xmark & S \\
\midrule
\rowcolor{gray!10}
\textbf{\benchmark(Ours)} & 270 & 540 Vs (273,068 Is) & \textbf{50+} & 2,932 & \cmark & \cmark & \cmark & S \\
\bottomrule
\end{tabular}
}
\label{tab:benchmark_compare}
\end{table*}

\section{\benchmark}

\begin{figure*}[h]
  \centering
    \includegraphics[width=\linewidth]{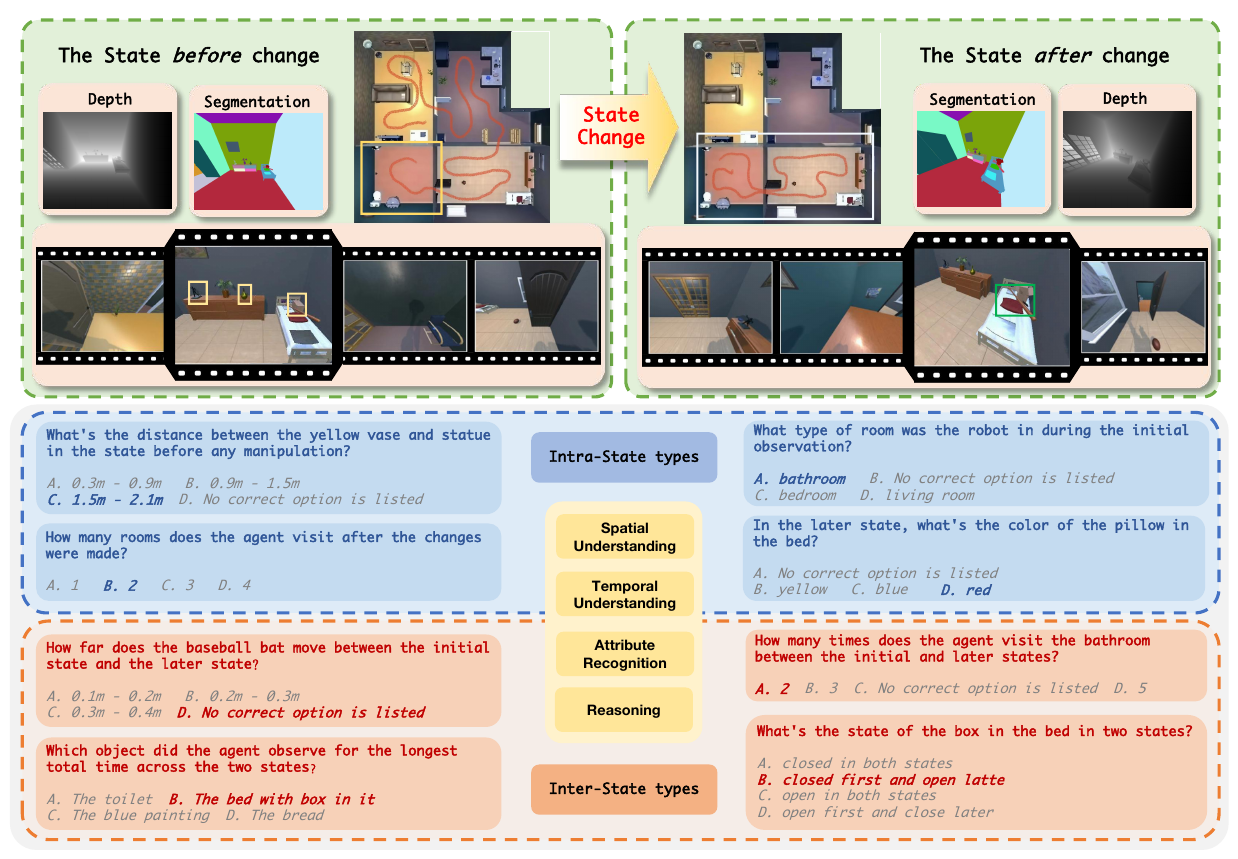}
    \caption{Overview of \textbf{\benchmark}. \textbf{Intra-State}~questions rely on a single state, while \textbf{Inter-State}~questions require comparing both \textit{``before''} and \textit{``after''} states. These questions collectively probe four core capabilities: \textbf{Spatial Understanding}, \textbf{Temporal Understanding}, \textbf{Attribute Recognition} and \textbf{Reasoning}.}
    \label{fig:overview}
\end{figure*}

\subsection{Overview}
We introduce \textbf{\benchmark}, a \textbf{M}ulti-modal, \textbf{M}ulti-state, \textbf{M}ulti-dimensional benchmark for evaluating the ability of Large Multimodal Models (LMMs) to reason about state transitions, which consists of two distinct subsets. 
\textbf{\simsubset}~is built upon the AI2-THOR simulator~\cite{AI2-THOR} and ProcTHOR-10k~\cite{ProcTHOR}, while \textbf{\realsubset}~is composed of real-world scenarios. Across both subsets, we establish paired states (\textit{``before''} and \textit{``after''}) defined by modifications to object presence, position, or internal attributes like being open/closed or toggled on/off. 

\begin{figure*}[t]
    \centering
    \begin{minipage}[c]{0.6\linewidth}
        \centering
        \includegraphics[width=\linewidth, height=7cm, keepaspectratio]{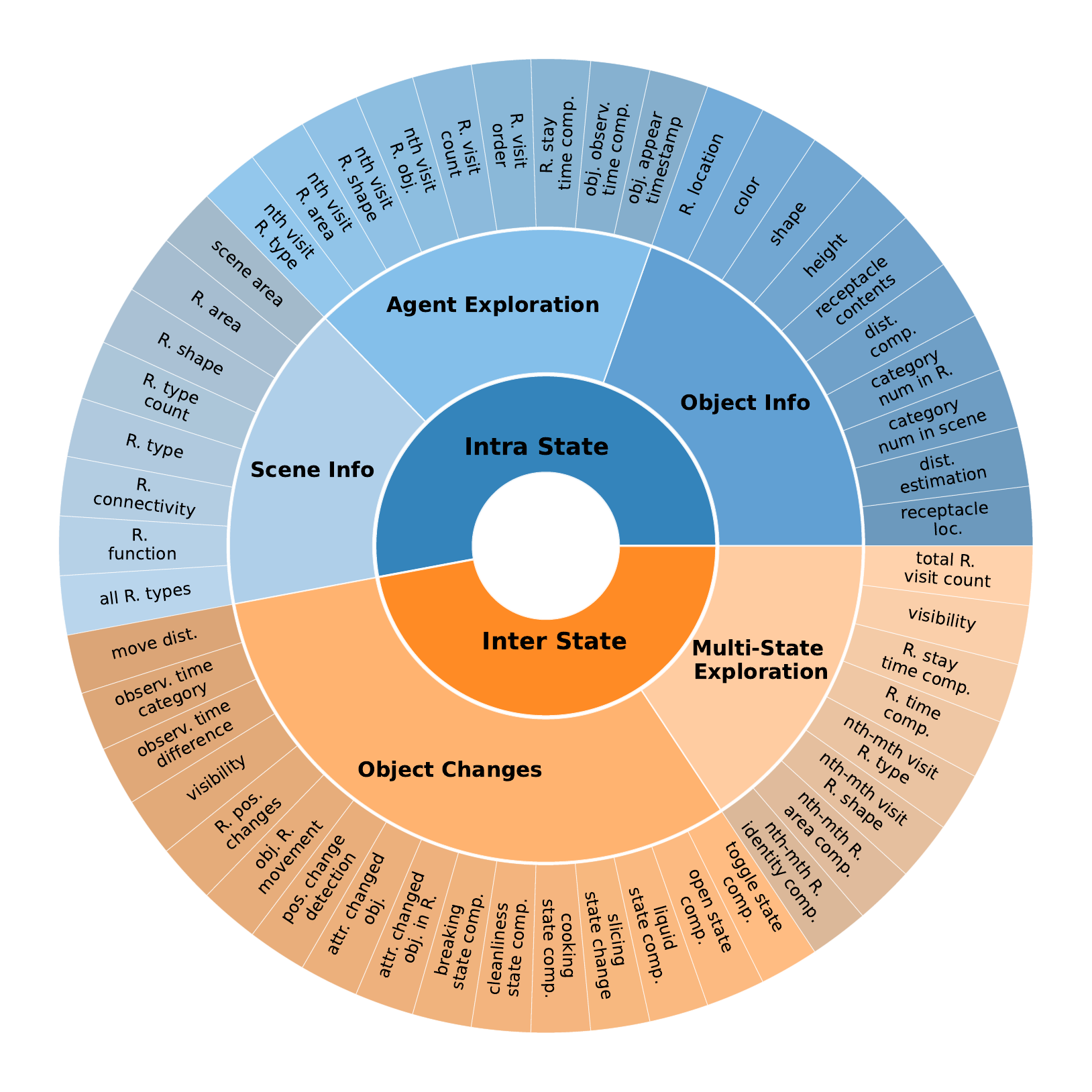} 
        \subcaption{Task categories in \textbf{\benchmark}.}
        \label{fig:subtasks}
    \end{minipage}%
    \hfill
    \begin{minipage}[c]{0.4\linewidth}
        \centering
        \includegraphics[width=\linewidth, height=7cm, keepaspectratio]{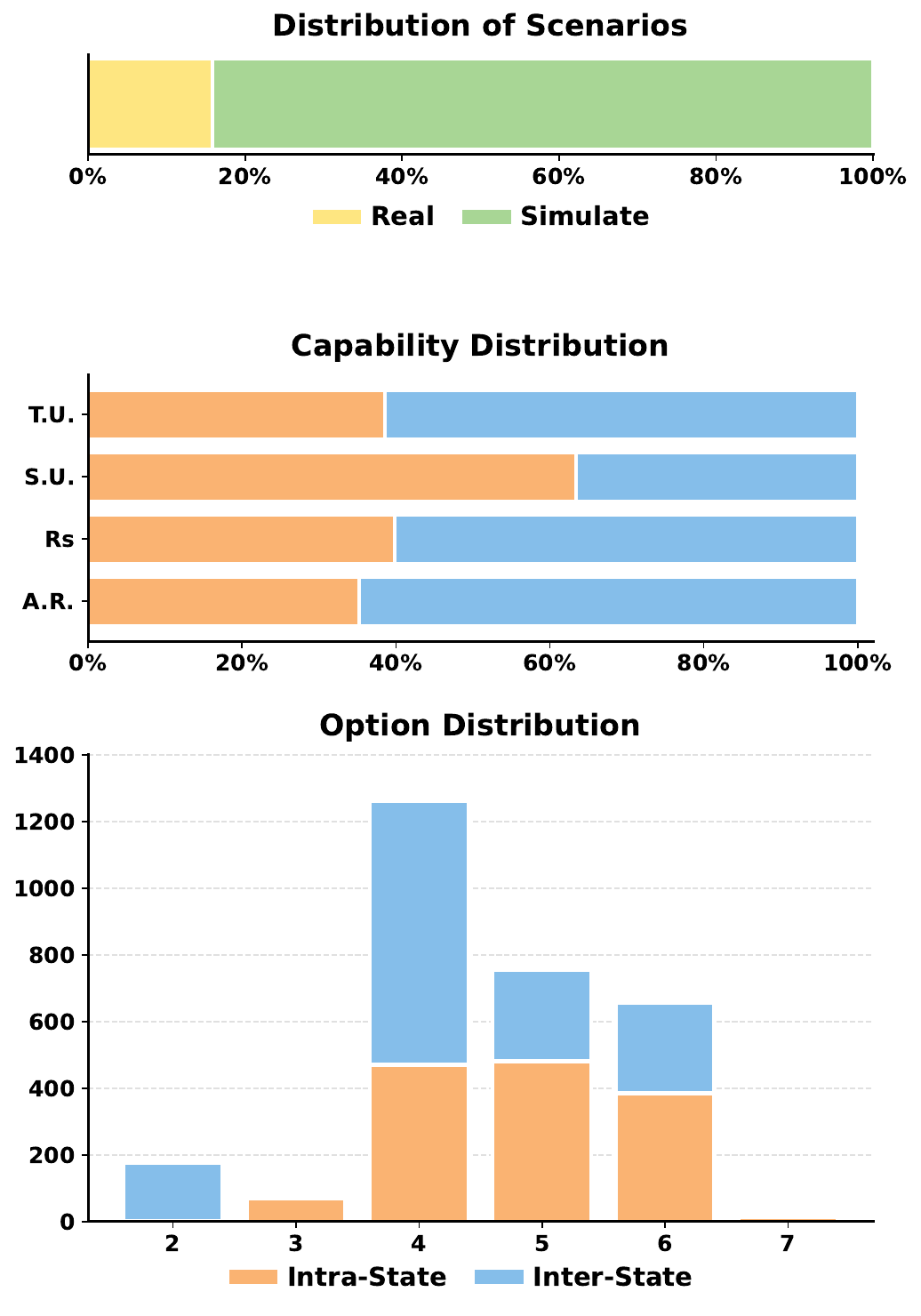} 
        \subcaption{Comprehensive distributions.}
        \label{fig:distributions}
    \end{minipage}
    
    \caption{\textbf{\benchmark~Statistic}. (a) \textbf{\benchmark} are classifiable into five primary categories according to the task, which are then further subdivided into more than 50 types of subtasks. (b) \textbf{Comprehensive Distributions:} This panel details the dataset's composition across three dimensions. \textit{Top:} The proportion of factual versus hallucination questions, designed to strictly evaluate model robustness. \textit{Middle:} The capability distribution across state types, showing that single questions evaluate multiple abilities (T.U., S.U., Rs, A.R.) in a compound manner. \textit{Bottom:} The distribution of candidate option counts. We adopt a variable number of answer options (from 2 to 7) to avoid the bias of fixed choices.}
    \label{fig:statistic}
\end{figure*}

As illustrated in \cref{fig:subtasks}, questions are divided into two main categories: \textbf{Intra-State} questions, answerable from a single state, and \textbf{Inter-State} questions, which require comparing two states to identify changes. To provide a comprehensive evaluation, these categories are further structured to probe \textbf{4 core capabilities}: \textit{Spatial Understanding}, \textit{Temporal Understanding}, \textit{Attribute Recognition}, and \textit{Reasoning}. These four pillars are ultimately broken down into \textbf{over 50 fine-grained subtasks}. Furthermore, as shown in Fig.~\ref{fig:distributions}, to more closely align with real-world situations and specifically evaluate models' susceptibility to hallucination, we also designed \textbf{hallucination-type} questions by replacing the correct choice with the option \textit{``No correct option is listed.''}. These hallucination-type questions account for \textbf{20.6\%} of the entire benchmark.

\subsection{\simsubset~Generation}

\begin{figure*}[t]
  \centering
    \includegraphics[width=\linewidth]{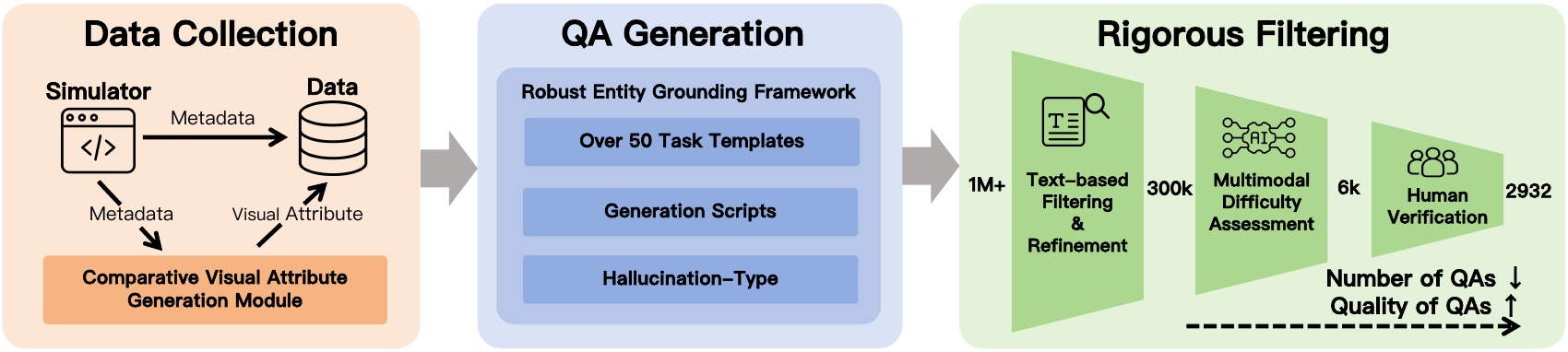}
    \caption{\textbf{\simsubset~generation pipeline.} The 3-stage process transforms raw simulation data into a high-quality benchmark through rigorous filtering.}
    \label{fig:generation}
\end{figure*}

\noindent\textbf{Data Collection and State Modification.}
We programmatically establish paired states for 270 indoor scenes. The state modifications include alterations to object presence, position, or internal attributes like being open/closed or toggled on/off, all while adhering to physical plausibility. For each state, we generate an exploration trajectory and capture egocentric videos, together with abundant backend metadata. Beyond serving as a rigorous evaluation tool, this rich aggregation of multi-modal metadata offers a versatile resource to catalyze future research in embodied AI and situated reasoning.


\noindent\textbf{Comparative Visual Attribute Generation Module.}
While the simulator provides basic object metadata, it critically lacks fine-grained visual attributes essential for human-like reasoning, such as color, shape, and texture. This deficiency becomes particularly problematic when distinguishing between multiple instances of the same category. To address this, we developed a \textbf{Comparative Visual Attribute Generation Module}. 
\begin{enumerate}
    \item We isolate all visual instances of each asset and select the \textbf{5 highest-quality observation views} via backend segmentation.
    \item The DAM~\cite{DAM} generates descriptions for these views, which are then synthesized by an LLM into a unified multi-view profile for each asset.
    \item The LLM performs a \textbf{comparative analysis} across assets of the same category to extract unique discriminative features (e.g., ``\textit{the black, rectangular table}'' vs. ``\textit{the round, white table}'').
\end{enumerate}

\noindent\textbf{Robust Entity Grounding Framework.}
Generating unambiguous questions for complex multi-object scenes is non-trivial, as referential ambiguity often arises when distinguishing between identical objects. To address this, we developed a Robust Entity Grounding Framework that dynamically constructs unique identifiers for every object. Rather than relying on generic simulator IDs, our framework fuses discriminative visual attributes with precise spatial contexts. For instance, it can uniquely identify \textit{``a red apple on the table''} versus \textit{``a green apple on the shelf''} by combining color, location, and relational information. This hierarchical grounding ensures that every question targets a specific entity, thereby eliminating confusion and forcing models to perform true visual localization rather than guessing based on category priors.

\noindent\textbf{Rigorous Filtering: Screening 1M Candidates to 3K High-Fidelity Samples.}
While existing VQA datasets often pursue massive scale, they are frequently plagued by language shortcuts or trivial correlations that allow models to answer correctly without looking at the visual content. Initially, our automated pipeline generated over \textbf{1 million} candidate QA pairs. However, empirical analysis revealed that such raw generations were riddled with severe issues: objects were often visually occluded, references remained ambiguous despite grounding efforts, and many state changes were too trivial to challenge modern models.

To construct a genuinely diagnostic testbed, we prioritized high density and ``anti-shortcut'' quality over sheer volume. We implemented a stringent, multi-stage filtration pipeline to screen this massive pool:

\begin{enumerate}
    \item \textbf{Text-based Filtering \& Refinement:} We first employ an LLM to process QA pairs without visual context. This includes a \textbf{Vision-free Answerability Check} to discard questions solvable by common sense or language priors alone, and \textbf{Ambiguity Removal} to eliminate subjective or confusing distinctions. Furthermore, we apply \textbf{Linguistic Enhancement} to rewrite template-based questions into natural, fluent, and human-like expressions.

    \item \textbf{Multimodal Difficulty Assessment:} To eliminate trivial samples, we filter candidates using a lightweight LMM~\cite{InternVL3.5}. Questions consistently solved across multiple trials are discarded as lacking sufficient challenge.

    \item \textbf{Intensive Human Verification:} Finally, a thorough human review process yielded the final set of \textbf{2,932} high-quality QA pairs.
\end{enumerate}

Ultimately, this rigorous selection process ensures that \textbf{\benchmark}~is free from superficial shortcuts, offering a highly credible and uncompromising evaluation of true multi-state reasoning capabilities.

\subsection{\realsubset~Generation}
This subset features 29 distinct real-world scenes ranging from household rooms to office spaces, all newly captured specifically for this benchmark to ensure zero overlap with existing public datasets and completely avoid potential data contamination. In each scene, human annotators physically manipulated objects to create distinct \textit{``before''} and \textit{``after''} states, meticulously documenting the specific changes for each object. Based on these documented state transitions, expert annotators manually crafted an initial pool of question-answer pairs. To ensure this subset serves as a rigorous upper-bound evaluation, we employed LMMs to filter these questions, retaining only the most challenging samples where models frequently failed despite clear visual evidence. This rigorous curation process yielded a final set of \textbf{552} highly-challenging questions.
\section{Evaluations}
\subsection{Task and Experimental Setup}

\paragraph{\textbf{Task Formulation.}}
To fully validate LMMs' ability to perceive scenes with changing states, our experiment involves inputting videos of a scene's two states into the LMM, along with a question and candidate options. When the videos are input, they are accompanied by textual descriptions to distinguish which state each represents. The LMM is then required to select the correct options.

\paragraph{\textbf{Models and Baselines.}}
We evaluate 22 Models and 2 baselines, organized into four distinct categories:
\begin{itemize}
    \item \textbf{2 Proprietary LMMs,} \texttt{Gemini-2.5-pro} (\textit{thinking\_budget=128}) and \texttt{GPT-5} (\textit{thinking effort=low}), both under minimal reasoning settings.
    
    \item \textbf{14 Open-Source LMMs,} with model sizes ranging from 2B to 235B, which can be further split into general LMMs (\textit{e.g.}, \texttt{Qwen3-VL}~\cite{qwen3-vl}), video-enhanced LMMs (\textit{e.g.}, \texttt{InternVideo2.5}~\cite{InternVideo2.5} and \texttt{Video-XL-2}~\cite{Video-XL-2}), and reasoning-enhanced LMMs (\textit{e.g.}, \texttt{WeThink}~\cite{WeThink}).
    
    \item \textbf{6 Vision Blind LLMs,} following~\cite{videommlu}, we test text-only LLM~\cite{qwen3} at 6 different scales without enabling the thinking mode.
    
    \item \textbf{2 Baselines,} we include \texttt{random} and \texttt{human (10 annotators with at least a bachelor's degree)} performance baselines.
\end{itemize}

\begin{table*}[t]
\centering
\renewcommand{\arraystretch}{1.08}
\small
\caption{Performance of various models on \textbf{\benchmark}. S. U. = Spatial Understanding, T. U. = Temporal Understanding, A. R. = Attribute Recognition, Rs = Reasoning. By default, models are evaluated with \textbf{200 input frames per video} at a resolution of \textbf{448$\times$448}, with thinking mode disabled. Models marked with \textsuperscript{*} use only 60 frames, and the model marked with \textsuperscript{\dag} employs an input resolution of 336$\times$448. The model marked with \textsuperscript{\S} is evaluated only on \simsubset.}
\resizebox{\textwidth}{!}{%
\begin{tabular}{l|c|c|c|cccc|c|cccc}
\toprule
   &
   &
   &
  \multicolumn{5}{c|}{\textbf{Intra-state}} &
  \multicolumn{5}{c}{\textbf{Inter-state}} \\   \cmidrule(lr){4-8} \cmidrule(lr){9-13}
  \multirow{-2}{*}{\textbf{Model}} &
  \multirow{-2}{*}{\textbf{Size}} &
  \multirow{-2}{*}{\textbf{Overall}} &
  \textbf{Avg.} &
  \textbf{S. U.} &
  \textbf{T. U.} &
  \textbf{A. R.} &
  \multicolumn{1}{c|}{\textbf{Rs}} &
  \textbf{Avg.} &
  \textbf{S. U.} &
  \textbf{T. U.} &
  \textbf{A. R.} &
  \textbf{Rs} \\ \midrule
  Human\textsuperscript{\S} &
  -- &
  \cellcolor[HTML]{A9A9A9}\color{white}89.75 &
  \cellcolor[HTML]{A9A9A9}\color{white}90.42 &
  \cellcolor[HTML]{A9A9A9}\color{white}83.39 &
  \cellcolor[HTML]{A9A9A9}\color{white}91.98 &
  \cellcolor[HTML]{A9A9A9}\color{white}90.89 &
  \cellcolor[HTML]{A9A9A9}\color{white}87.91 &
  \cellcolor[HTML]{A9A9A9}\color{white}89.29 &
  \cellcolor[HTML]{A9A9A9}\color{white}88.26 &
  \cellcolor[HTML]{A9A9A9}\color{white}90.06 &
  \cellcolor[HTML]{A9A9A9}\color{white}92.25 &
  \cellcolor[HTML]{A9A9A9}\color{white}88.79 \\
  Random &
  -- &
  \cellcolor[HTML]{F2F2F2}24.14 &
  \cellcolor[HTML]{F2F2F2}22.52 &
  \cellcolor[HTML]{F2F2F2}21.76 &
  \cellcolor[HTML]{F2F2F2}22.61 &
  \cellcolor[HTML]{F2F2F2}22.39 &
  \cellcolor[HTML]{F2F2F2}22.67 &
  \cellcolor[HTML]{F2F2F2}25.66 &
  \cellcolor[HTML]{F2F2F2}24.72 &
  \cellcolor[HTML]{F2F2F2}26.27 &
  \cellcolor[HTML]{F2F2F2}22.93 &
  \cellcolor[HTML]{F2F2F2}25.54 \\ \midrule
\multicolumn{13}{l}{\textit{Vision-blind Models}} \\ 
\multirow{6}{*}{Qwen3\cite{qwen3}}
&  0.6B & 20.42  & 22.09  & 22.29 & 26.80  & 21.80 & 21.67 & 19.15  & 20.24  & 19.45 & 20.19 & 14.46  \\
& 1.4B & 24.45 & 23.92 & 27.43  & 27.24  & 26.00 & 18.89 & 24.93 & 27.59  & 23.47  & 22.39 & 26.93  \\
& 4B  & 28.78 & 31.04  & 30.78  & 26.30 & 33.57  & 32.91 & 26.74 & 30.24 & 26.26 & 23.54  & 26.80   \\
& 8B   & 32.22  & 32.60 & 29.72  & 35.82  & 33.66 & 38.35 & 32.40 & 32.43  & 30.98  & 30.68 & 38.40 \\
& 14B  & 34.37  & 33.15  & 33.33  & 38.52 & 33.06  & 34.83 & 34.79  & 36.99  & 33.83 & 33.38 & 37.24  \\
& 32B  & 32.97 & 34.37 & 33.92  & 41.66  & 34.58 & 36.52  & 30.82  & 34.62 & 29.43  & 29.21 & 30.75 \\ \midrule
\multicolumn{13}{l}{\textit{Proprietary LMMs}} \\ 
  Gemini-2.5-pro &
  -- &
  \cellcolor[HTML]{94C9BF}47.58 &
  \cellcolor[HTML]{80C0B3}53.21 &
  \cellcolor[HTML]{79BCAF}55.39 &
  \cellcolor[HTML]{9ECEC5}44.65 &
  \cellcolor[HTML]{6FB7A9}58.25 &
  \cellcolor[HTML]{8FC7BC}48.83 &
  \cellcolor[HTML]{A8D3CB}41.83 &
  \cellcolor[HTML]{9DCEC4}44.91 &
  \cellcolor[HTML]{A7D3CA}42.03 &
  \cellcolor[HTML]{A0D0C7}43.88 &
  \cellcolor[HTML]{BBDDD7}36.08 \\
  GPT-5\textsuperscript{\S} &
  -- &
  \cellcolor[HTML]{9CCEC4}45.14 &
  \cellcolor[HTML]{9CCDC4}45.20 &
  \cellcolor[HTML]{96CBC0}46.82 &
  \cellcolor[HTML]{97CBC1}46.58 &
  \cellcolor[HTML]{87C3B7}51.20 &
  \cellcolor[HTML]{B4D9D2}38.27 &
  \cellcolor[HTML]{9CCDC4}45.29 &
  \cellcolor[HTML]{96CBC1}46.81 &
  \cellcolor[HTML]{97CBC1}46.65 &
  \cellcolor[HTML]{97CBC1}46.51 &
  \cellcolor[HTML]{A6D2CA}42.37 \\ \midrule
\multicolumn{13}{l}{\textit{Open-Source LMMs}} \\ 
  WeThink-VL-7B~\cite{WeThink} &
  7B &
  \cellcolor[HTML]{EFF7F5}21.03 &
  \cellcolor[HTML]{EDF6F4}21.59 &
  \cellcolor[HTML]{F0F7F6}20.89 &
  \cellcolor[HTML]{FFFFFF}16.39 &
  \cellcolor[HTML]{E3F1EF}24.41 &
  \cellcolor[HTML]{ECF5F3}22.04 &
  \cellcolor[HTML]{F0F7F6}20.80 &
  \cellcolor[HTML]{EAF4F2}22.59 &
  \cellcolor[HTML]{F2F8F7}20.21 &
  \cellcolor[HTML]{F2F8F7}20.17 &
  \cellcolor[HTML]{EBF5F3}22.33 \\
  Video-R1-7B~\cite{Video-R1} &
  7B &
  \cellcolor[HTML]{C4E2DC}33.40 &
  \cellcolor[HTML]{C2E0DA}34.20 &
  \cellcolor[HTML]{BFDFD9}34.90 &
  \cellcolor[HTML]{B6DAD3}37.71 &
  \cellcolor[HTML]{C1E0DA}34.41 &
  \cellcolor[HTML]{C3E1DB}33.79 &
  \cellcolor[HTML]{C6E2DD}33.00 &
  \cellcolor[HTML]{C3E1DB}33.72 &
  \cellcolor[HTML]{C3E1DB}33.84 &
  \cellcolor[HTML]{CCE6E1}31.08 &
  \cellcolor[HTML]{C6E2DD}33.07 \\
  Video-XL-2~\cite{Video-XL-2} &
  8B &
  \cellcolor[HTML]{C6E3DD}32.86 &
  \cellcolor[HTML]{C1E0DA}34.50 &
  \cellcolor[HTML]{BFDFD9}35.07 &
  \cellcolor[HTML]{AFD7CF}39.79 &
  \cellcolor[HTML]{C2E0DA}34.19 &
  \cellcolor[HTML]{C3E1DB}33.81 &
  \cellcolor[HTML]{C9E4DF}32.10 &
  \cellcolor[HTML]{C6E3DD}32.95 &
  \cellcolor[HTML]{C7E3DD}32.81 &
  \cellcolor[HTML]{CFE7E3}30.23 &
  \cellcolor[HTML]{CAE5E0}31.71 \\
  MiniCPM-V-4.5~\cite{MiniCPM-V-4.5} &
  8B &
  \cellcolor[HTML]{B9DCD5}36.65 &
  \cellcolor[HTML]{AAD5CC}41.01 &
  \cellcolor[HTML]{A9D4CC}41.37 &
  \cellcolor[HTML]{8AC4B9}50.54 &
  \cellcolor[HTML]{ACD5CD}40.54 &
  \cellcolor[HTML]{B2D8D1}38.81 &
  \cellcolor[HTML]{C6E2DD}33.07 &
  \cellcolor[HTML]{C9E4DF}32.17 &
  \cellcolor[HTML]{BCDED7}35.73 &
  \cellcolor[HTML]{CCE6E1}31.10 &
  \cellcolor[HTML]{CDE6E1}30.97 \\
  InternVideo2.5-8B\textsuperscript{*}~\cite{InternVideo2.5} &
  8B &
  \cellcolor[HTML]{BEDFD8}35.17 &
  \cellcolor[HTML]{ACD5CD}40.60 &
  \cellcolor[HTML]{A6D2CA}42.29 &
  \cellcolor[HTML]{A9D4CC}41.38 &
  \cellcolor[HTML]{A2D1C7}43.41 &
  \cellcolor[HTML]{BCDED7}35.74 &
  \cellcolor[HTML]{CEE7E2}30.57 &
  \cellcolor[HTML]{CBE5E0}31.48 &
  \cellcolor[HTML]{CDE6E1}30.91 &
  \cellcolor[HTML]{CCE6E1}31.12 &
  \cellcolor[HTML]{D7EBE7}28.11 \\
  VideoLLaMA3-7B\textsuperscript{\dag,*}\cite{VideoLLaMA3} &
  7B &
  \cellcolor[HTML]{B1D8D1}38.98 &
  \cellcolor[HTML]{AAD4CC}41.21 &
  \cellcolor[HTML]{ABD5CD}40.72 &
  \cellcolor[HTML]{9ACCC2}45.89 &
  \cellcolor[HTML]{A2D1C8}43.36 &
  \cellcolor[HTML]{B3D9D2}38.44 &
  \cellcolor[HTML]{B8DBD4}37.13 &
  \cellcolor[HTML]{BDDED8}35.46 &
  \cellcolor[HTML]{B4DAD2}38.12 &
  \cellcolor[HTML]{ACD5CD}40.53 &
  \cellcolor[HTML]{C3E1DB}33.72 \\
  LLaVA-OneVision-1.5-8B~\cite{Llava-OneVision-1.5} &
  8B &
  \cellcolor[HTML]{D0E8E3}29.91 &
  \cellcolor[HTML]{D1E8E3}29.90 &
  \cellcolor[HTML]{CBE5E0}31.53 &
  \cellcolor[HTML]{D1E8E4}29.75 &
  \cellcolor[HTML]{CBE5E0}31.53 &
  \cellcolor[HTML]{DFEFEC}25.64 &
  \cellcolor[HTML]{D0E7E3}30.06 &
  \cellcolor[HTML]{DDEEEB}26.35 &
  \cellcolor[HTML]{CEE7E2}30.59 &
  \cellcolor[HTML]{BEDED8}35.32 &
  \cellcolor[HTML]{D6EBE7}28.24 \\
  InternVL3.5-2B\textsuperscript{*}~\cite{InternVL3.5} &
  2B &
  \cellcolor[HTML]{C0DFD9}34.72 &
  \cellcolor[HTML]{C0E0DA}34.60 &
  \cellcolor[HTML]{BBDDD6}36.22 &
  \cellcolor[HTML]{B6DBD4}37.53 &
  \cellcolor[HTML]{C3E1DB}33.87 &
  \cellcolor[HTML]{C4E1DC}33.59 &
  \cellcolor[HTML]{BDDED8}35.48 &
  \cellcolor[HTML]{B8DCD5}36.91 &
  \cellcolor[HTML]{C5E2DC}33.21 &
  \cellcolor[HTML]{B2D9D1}38.67 &
  \cellcolor[HTML]{B8DBD4}37.10 \\
  InternVL3.5-4B\textsuperscript{*}~\cite{InternVL3.5} &
  4B &
  \cellcolor[HTML]{BBDDD6}36.22 &
  \cellcolor[HTML]{B7DBD4}37.23 &
  \cellcolor[HTML]{BADDD6}36.38 &
  \cellcolor[HTML]{BFDFD9}34.88 &
  \cellcolor[HTML]{B5DAD3}37.88 &
  \cellcolor[HTML]{B2D9D1}38.76 &
  \cellcolor[HTML]{C0DFD9}34.73 &
  \cellcolor[HTML]{BCDED7}35.84 &
  \cellcolor[HTML]{C7E3DD}32.79 &
  \cellcolor[HTML]{B5DAD3}38.03 &
  \cellcolor[HTML]{C1E0DA}34.52 \\
  InternVL3.5-8B\textsuperscript{*}~\cite{InternVL3.5} &
  8B &
  \cellcolor[HTML]{C0DFD9}34.82 &
  \cellcolor[HTML]{ABD5CD}40.76 &
  \cellcolor[HTML]{ABD5CD}40.82 &
  \cellcolor[HTML]{C8E3DE}32.51 &
  \cellcolor[HTML]{ABD5CD}40.93 &
  \cellcolor[HTML]{A7D3CA}41.99 &
  \cellcolor[HTML]{CEE6E2}30.69 &
  \cellcolor[HTML]{D1E8E3}29.86 &
  \cellcolor[HTML]{C5E2DD}33.18 &
  \cellcolor[HTML]{D3E9E5}29.09 &
  \cellcolor[HTML]{D3E9E5}29.04 \\
  Qwen3-VL-2B~\cite{qwen3-vl} &
  2B &
  \cellcolor[HTML]{BFDFD9}34.90 &
  \cellcolor[HTML]{AED6CF}39.94 &
  \cellcolor[HTML]{A5D2C9}42.59 &
  \cellcolor[HTML]{ACD5CD}40.62 &
  \cellcolor[HTML]{B4D9D2}38.20 &
  \cellcolor[HTML]{B0D8D0}39.35 &
  \cellcolor[HTML]{CEE7E2}30.57 &
  \cellcolor[HTML]{CEE7E2}30.61 &
  \cellcolor[HTML]{D2E8E4}29.54 &
  \cellcolor[HTML]{CDE6E1}30.82 &
  \cellcolor[HTML]{C3E1DB}33.84 \\
  Qwen3-VL-4B~\cite{qwen3-vl} &
  4B &
  \cellcolor[HTML]{B2D8D1}38.79 &
  \cellcolor[HTML]{ACD5CD}40.56 &
  \cellcolor[HTML]{A4D1C9}42.90 &
  \cellcolor[HTML]{B6DAD3}37.70 &
  \cellcolor[HTML]{A8D3CB}41.72 &
  \cellcolor[HTML]{BEDED8}35.32 &
  \cellcolor[HTML]{BDDED8}35.58 &
  \cellcolor[HTML]{BCDED7}35.78 &
  \cellcolor[HTML]{BCDED7}35.74 &
  \cellcolor[HTML]{AED7CF}39.83 &
  \cellcolor[HTML]{C7E3DE}32.56 \\
  Qwen3-VL-8B~\cite{qwen3-vl} &
  8B &
  \cellcolor[HTML]{B1D8D0}39.07 &
  \cellcolor[HTML]{9ECFC5}44.50 &
  \cellcolor[HTML]{A4D1C8}42.96 &
  \cellcolor[HTML]{A7D3CA}42.01 &
  \cellcolor[HTML]{95CAC0}47.20 &
  \cellcolor[HTML]{A1D0C7}43.86 &
  \cellcolor[HTML]{C3E1DB}33.93 &
  \cellcolor[HTML]{C5E2DC}33.30 &
  \cellcolor[HTML]{BEDFD8}35.26 &
  \cellcolor[HTML]{C3E1DB}33.83 &
  \cellcolor[HTML]{C5E2DD}33.15 \\
  Qwen3-VL-235B-A22B\textsuperscript{\S}~\cite{qwen3-vl} &
  235B &
  \cellcolor[HTML]{B3D9D1}38.60 &
  \cellcolor[HTML]{BADCD6}36.46 &
  \cellcolor[HTML]{B6DAD3}37.74 &
  \cellcolor[HTML]{B5DAD3}37.88 &
  \cellcolor[HTML]{B0D7D0}39.38 &
  \cellcolor[HTML]{CAE4DF}31.82 &
  \cellcolor[HTML]{AAD4CC}41.13 &
  \cellcolor[HTML]{B0D7D0}39.40 &
  \cellcolor[HTML]{AED7CF}39.92 &
  \cellcolor[HTML]{9DCEC4}44.93 &
  \cellcolor[HTML]{A7D3CB}41.85 \\ \bottomrule
\end{tabular}
}
\label{tab:basemodel_exp}
\end{table*}



\subsection{Zero-shot Evaluation Results}
We present the main evaluation results in Tab.~\ref{tab:basemodel_exp}. The results are analyzed across different model families, scales, and capabilities. Our analysis reveals several key insights into the current state of LMMs for state change understanding:

\paragraph{\textbf{1) Human-Machine Performance Gap.}}
A central finding from our evaluation is that all models perform poorly on \textbf{\benchmark}. While humans comfortably achieve an \textbf{89.75\%} overall accuracy, demonstrating robust understanding of dynamic state changes, even the most advanced proprietary models struggle. \texttt{Gemini-2.5-pro} and \texttt{GPT-5} top the leaderboard with 47.58\% and 45.14\% respectively, yet still fall far short of human performance. Among open-source models, the vast majority of models languish between the 20\% and 35\% range. This stark contrast underscores that fine-grained comparative reasoning remains a formidable challenge for today's AI systems.

\paragraph{\textbf{2) The Critical Role of Vision and Non-Monotonic Scaling.}}
The results confirm that visual input is essential for the benchmark, as vision-enabled LMMs consistently outperform their vision-blind LLM counterparts of similar sizes. Simultaneously, the performance within the vision-blind \texttt{Qwen3}~\cite{qwen3} family shows a clear positive correlation with model size, suggesting that a more powerful language backbone provides a better foundation for multimodal tasks. More details will be shown in the Appendix.

\begin{figure*}[t]
  \centering
    \includegraphics[width=\linewidth]{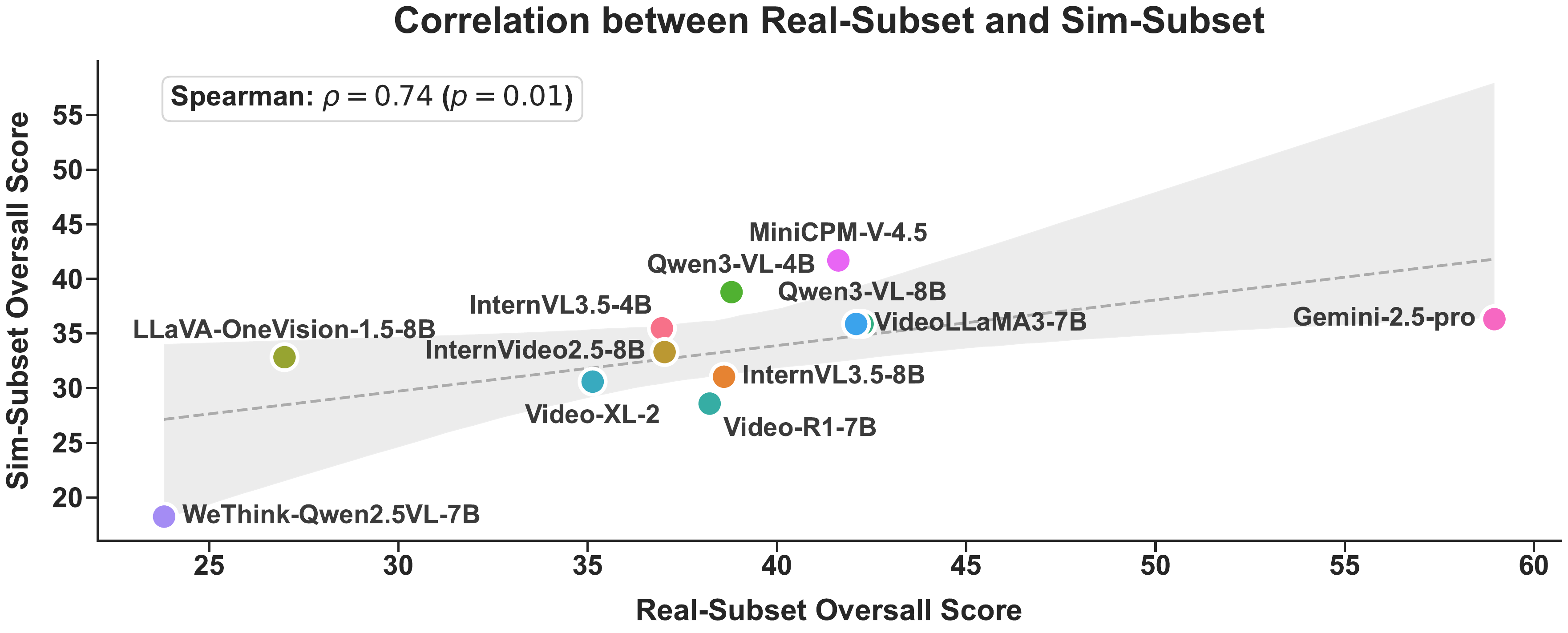}
    \caption{\textbf{Correlation Analysis between \simsubset~and \realsubset.} 
    Each point represents a model's performance on both subsets. The strong positive correlation (Pearson $r=0.74$) demonstrates that \textbf{\simsubset}~serves as a reliable diagnostic proxy for evaluating state-change reasoning capabilities, effectively predicting relative model rankings in real-world scenarios.}
    \label{fig:correlation}
\end{figure*}

\paragraph{\textbf{3) Validity of Simulation as a Diagnostic Proxy.}}
A fundamental challenge in synthetic benchmarking is ensuring alignment with real-world model capabilities. To validate the quality of our data construction, we analyzed the performance alignment across both \textbf{\simsubset} and \textbf{\realsubset}. As illustrated in Fig.~\ref{fig:correlation}, the results exhibit a statistically significant positive linear correlation, achieving a Spearman rank coefficient of $\rho=0.74$ ($p=0.01$). This alignment confirms that despite the domain gap caused by factors like texture differences and camera motion, the core reasoning challenges embedded in our simulated tasks remain structurally consistent with physical reality. Consequently, \textbf{\simsubset}~serves as a high-quality, scalable proxy for evaluating state-change reasoning, effectively predicting relative capability rankings in the real world while offering finer-grained diagnostic granularity.

\paragraph{\textbf{4) The Inherent Challenge of Inter-State Tracking.}}
A comparison between Intra-state and Inter-state understanding reveals the inherent difficulty of tracking dynamic transitions. Overall, the evaluated models generally exhibit lower performance on Inter-state tasks compared to their Intra-state counterparts. This discrepancy aligns with the cognitive intuition that comparing discrete temporal observations demands a higher level of spatial-temporal reasoning than merely parsing a static scene. Notably, this performance gap becomes significantly more pronounced when evaluating on \textbf{\realsubset}, where models must grapple with the added complexities of natural physical perturbations. The stark contrast between Intra-state and Inter-state accuracy in these physical scenarios highlights a critical vulnerability in current LMMs' ability to maintain robust semantic alignment across long temporal contexts. A detailed breakdown of performance differences across the two subsets is provided in the Appendix.

\paragraph{\textbf{5) The Formidable Challenge of Complex Reasoning.}}
Across both intra-state and inter-state evaluation, a clear limitation in models' cognitive capabilities emerges. Regardless of whether they are proprietary models or open-source variants, \textit{Reasoning (Rs)} consistently proves to be the most formidable challenge. In all evaluated LMMs, the Reasoning subtask yields lower performance compared to perception-based queries, highlighting a universal bottleneck. This pattern indicates that current LMMs remain fundamentally inadequate when tasked with multi-step logical operations, counting, or deductive spatial reasoning, whether within a single scene or across dynamic transitions.

\paragraph{\textbf{6) Robust Instruction Following as a Prerequisite.}}
While complex reasoning is a universal bottleneck, models specifically designed to enhance these capabilities do not necessarily provide a solution. The performance of \texttt{WeThink}~\cite{WeThink} (21.03\% overall) is particularly revealing. Despite being explicitly trained for complex reasoning, its output logs demonstrate a failure in basic instruction following, as it consistently failed to generate the required number of options specified by the prompt. This fundamental format violation, rather than an inherent flaw in its reasoning logic, led to its low accuracy. This phenomenon underscores that advanced reasoning models can suffer from a degradation in foundational instruction-following abilities. Consequently, robust instruction following remains a mandatory prerequisite for effectively deploying and evaluating complex multimodal reasoning in practice.

\begin{figure}[t]
  \centering
  \begin{subfigure}{0.50\linewidth}
    \includegraphics[width=\textwidth]{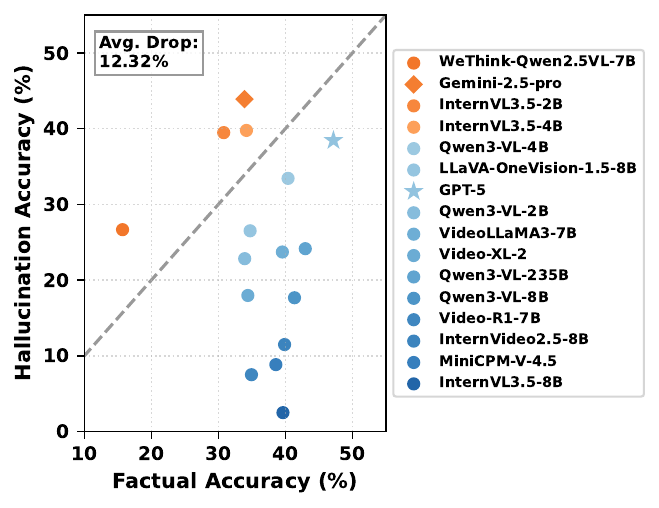}
    \caption{\textbf{Hallucination Analysis.} Most models suffer from severe hallucination issues. Darker blue shades indicate more severe hallucination.}
    \label{fig:hallucination}
  \end{subfigure}
  \hfill
  \begin{subfigure}{0.48\linewidth}
    \includegraphics[width=\textwidth]{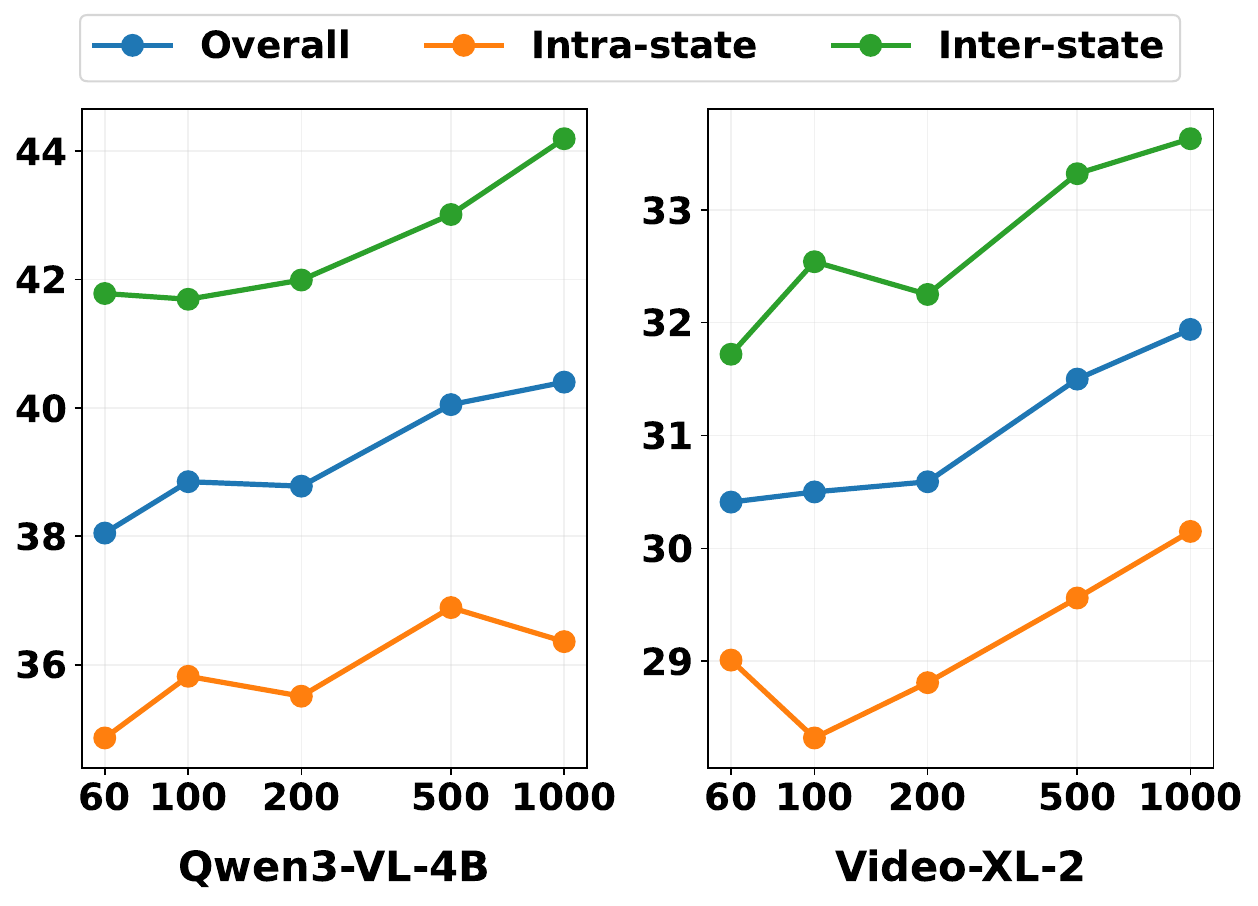}
    \caption{\textbf{Impact of Frame Sampling.} Performance trends for Qwen3-VL-4B-Instruct~\cite{qwen3-vl} and Video-XL-2~\cite{Video-XL-2} as input frames increase from 60 to 1000, showing a clear benefit from denser temporal information.}
    \label{fig:frame_sampling}
  \end{subfigure}
  \caption{
    \textbf{Diagnostic Analysis of Model Robustness and Temporal Sensitivity.} 
    \textbf{a)} A comparison of model performance on factual versus hallucination-type questions reveals an average accuracy drop of 12.32\%, highlighting a pervasive weakness in rejecting incorrect options. \textbf{(b)} An ablation study on frame density demonstrates that increasing the number of sampled frames consistently improves performance, particularly for Inter-state reasoning tasks.
  }
  \label{fig:anaysis}
\end{figure}

\paragraph{\textbf{7) The Pervasive Challenge of Hallucination.}}
A critical finding from our evaluation is the widespread difficulty models face with hallucination-type questions. As shown in Fig.~\ref{fig:hallucination}, most LMMs perform substantially worse when tasked with questions designed to elicit hallucinatory responses compared to factual ones. On average, there is a performance drop of 12.32 points on hallucination-centric questions relative to their fact-based counterparts. This pronounced weakness in handling potential hallucinations is a primary contributor to the modest overall performance scores, indicating a critical area for future model development. More Details will be shown in Appendix.

\paragraph{\textbf{8) Impact of Frame Sampling on Performance.}}
To verify the impact of the number of sampled frames on the test results, we conducted tests on \texttt{Qwen3-VL-4B-Instruct}~\cite{qwen3-vl} and \texttt{Video-XL-2}~\cite{Video-XL-2} with different numbers of sampled frames. As illustrated in Fig.~\ref{fig:anaysis}(b), within the range of sampling up to at most 1000 frames per video, the ``Overall'', ``Inter-state'' and ``Intra-state'' scores all present an upward trend as the number of frames increases, underscoring the benefit of denser temporal sampling for understanding state changes.

\section{Diagnostic Analysis: Isolating the Bottleneck}

\begin{figure*}[t]
  \centering
  \begin{subfigure}{0.59\linewidth}
    \centerline{\includegraphics[width=\textwidth]{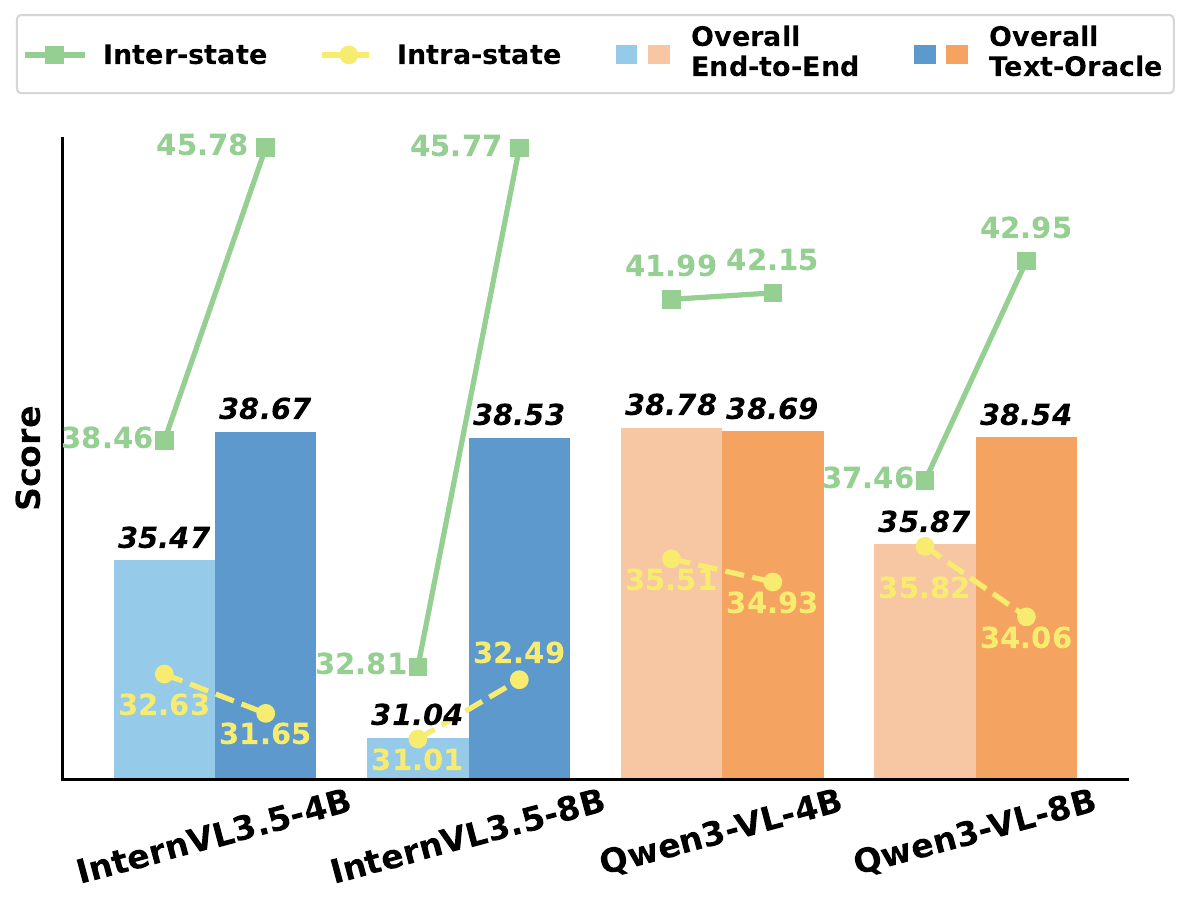}}
    \caption{End-to-End LMM vs. Text-Oracle Probe}
    \label{fig:baseline overall}
  \end{subfigure}
  \hfill
  \begin{subfigure}{0.40\linewidth}
    \centerline{\includegraphics[width=\textwidth]{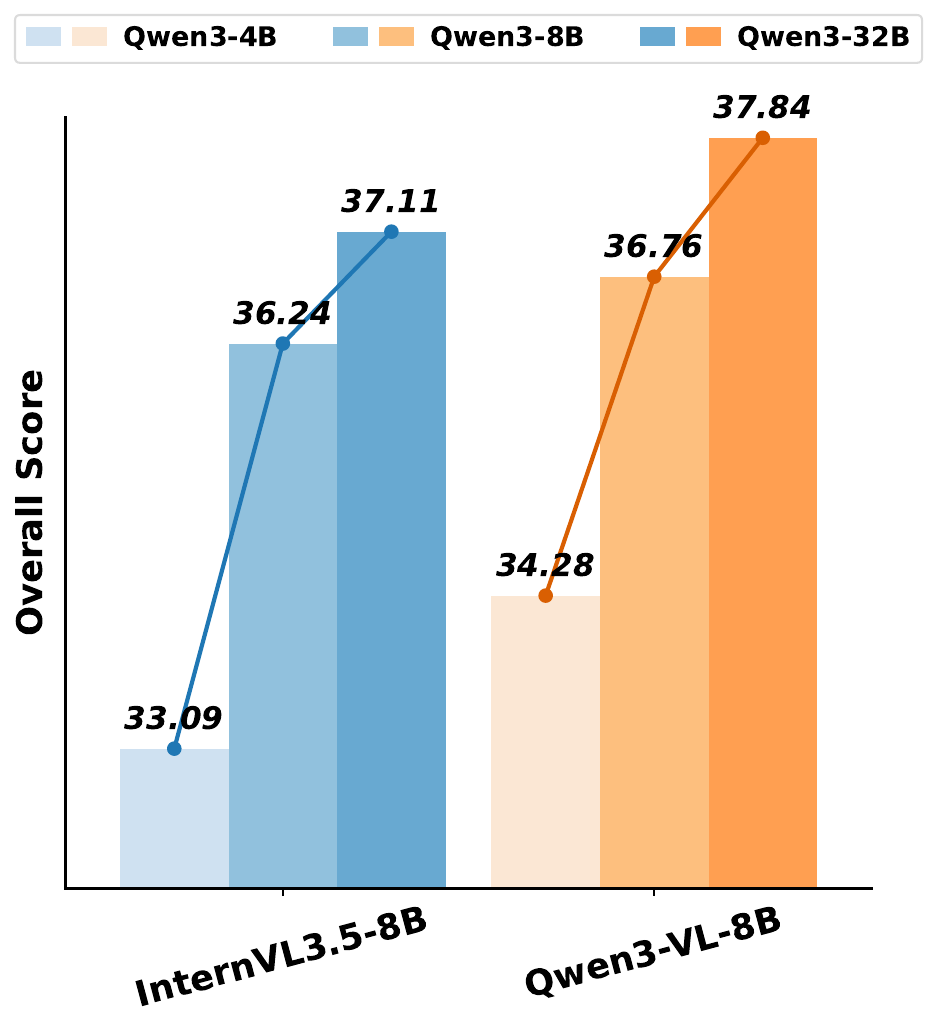}}
    \caption{Decoupled Textual LLM Scaling}
    \label{fig:baseline LLM}
  \end{subfigure}
  \caption{\textbf{Diagnostic Analysis of the Reasoning Bottleneck.} 
  (a) By explicitly serializing visual streams into text (Text-Oracle), the same LMM achieves significant performance gains over the standard End-to-End baseline. Notably, the improvements are highly concentrated on Inter-state tasks (green lines), isolating the bottleneck to cross-temporal visual tracking. 
  (b) When replacing the LMM's reasoning component with progressively larger external LLMs while keeping the text-oracle input, the overall performance steadily increases, demonstrating that the ultimate upper bound is dictated by logical reasoning capacity once visual dilution is resolved.}
  \label{fig:baseline}
\end{figure*}

\subsection{Text-Oracle Probe}
To address the inherent challenges LMMs face in processing long-form videos and tracking temporal state changes, we designed a diagnostic experimental setup, which we term the \textbf{Text-Oracle Probe}. This probe artificially bypasses the long-context visual processing phase by explicitly serializing the visual stream into text before reasoning. The probe operates through a 3-stage process:
\begin{enumerate}
    \item \textbf{Video Segmentation and Dense Captioning.} A given video is first partitioned into a sequence of non-overlapping temporal clips. An LMM is then applied to generate a detailed textual description of each clip along with its start time and end time. This converts the visual stream into a set of captions, each associated with a specific time interval.
    \item \textbf{Cohesive Summary Generation.} The sequence of time-coded captions is then consolidated by LLM or LMM, into a single, cohesive narrative summary that represents the entire video content.
    \item \textbf{Text-based Question Answering.} Finally, the generated summaries of two videos are concatenated with the question. An LLM or LMM is applied to answer the question based on text-only descriptions of the videos.
\end{enumerate}

\subsection{Diagnostic Results and Analysis}
We conducted this diagnostic experiment on \simsubset~with \texttt{InternVL3.5}~\cite{InternVL3.5} and \texttt{Qwen3-VL}~\cite{qwen3-vl} at 4B and 8B scale . As presented in Fig.~\ref{fig:baseline}, the results demonstrate a stark contrast between standard end-to-end performance and the text-oracle setting. This significant discrepancy exposes profound mechanistic flaws in how current LMMs handle long-term comparative reasoning.

\paragraph{\textbf{Performance Discrepancy under the Text-Oracle Setting.}}
Here we utilize the LMM to execute all stages of the probe: generating clip captions, summarizing them, and eventually answering the question based on the text. The results in Fig.~\ref{fig:baseline} reveal that this text-oracle setup consistently and significantly yields higher performance than the standard direct video QA baseline across most tested models. For instance, \texttt{InternVL3.5-8B-Instruct}~\cite{InternVL3.5} sees its overall accuracy jump by over 7 absolute points when evaluated through this text-only intermediate route. Similarly, \texttt{Qwen3-VL-8B-Instruct}~\cite{qwen3-vl} achieves an improvement of more than 4 absolute points compared to its standard end-to-end counterpart. The most pronounced discrepancies appear in the \textbf{``inter-state''} evaluation, which represents the core comparative challenge of our benchmark. For \texttt{InternVL3.5-8B-Instruct}~\cite{InternVL3.5}, a massive portion of initially incorrect predictions are successfully converted to correct answers under the text-oracle setting, causing the ``inter-state'' average score to skyrocket by nearly 9 absolute points. A similar leap of nearly 7 absolute points is observed for \texttt{InternVL3.5-4B-Instruct}~\cite{InternVL3.5}. Conversely, the performance on intra-state tasks remains relatively stable with minimal fluctuation.

\paragraph{\textbf{Analysis of Global Information Utilization.}}
This observation provides key insights into the internal workings of the model. That the same LMM achieves significantly higher accuracy simply by reasoning over its own generated captions demonstrates that the necessary answer information is successfully captured by the vision encoder. This implies that the intrinsic perceptual capacity of the visual encoder is not the primary bottleneck, but rather that this capacity is not fully utilized in the end-to-end setting. In the standard end-to-end process, question-relevant visual information is typically sparse, often appearing only in specific regions of a few frames. In contrast to the compact nature of text tokens, the number of visual tokens is vastly larger, resulting in a much lower information density for these critical cues. Consequently, these sparse visual tokens can easily become diluted or inaccessible amidst the massive influx of irrelevant background information over a long temporal window. However, when these visual cues are explicitly serialized into high-density text representations, the model successfully processes and retains them, as text is inherently more compatible with the processing mechanisms of the language backbone.


\paragraph{\textbf{Further Probe: LMM vs. Pure LLM for Textual Reasoning}}
To push our diagnosis further, we explored replacing the reasoning component of the Text-Oracle with external, text-only LLMs~\cite{qwen3}. As shown in Fig.~\ref{fig:baseline LLM}, our findings reveal a clear scaling trend: when using an external LLM of a comparable size to the LMM's language backbone, the decoupled system falls short of the integrated LMM operating in Text-Oracle mode. However, as we substitute progressively larger and more powerful text-only LLMs (e.g., 32B), the performance of the decoupled system steadily increases, eventually exceeding the limits of the integrated LMM. This diagnostic outcome yields two critical insights:
\begin{enumerate}
    \item \textbf{The value of integrated training:} The integrated LMM (when given text-oracle context) still outperforms a blindly decoupled LLM of similar size. This indicates that joint multimodal training preserves a richer, more contextually-aware representation even when reasoning is forced through text intermediate stages.
    
    \item \textbf{Reasoning ceiling dictates the upper bound:} The ultimate performance on this comparative task is heavily contingent on pure logical capacity. The fact that a massively scaled reasoning-focused LLM can compensate for and ultimately surpass the native LMM pipeline underscores that while visual alignment is the current primary bottleneck, advancing pure logical reasoning capability remains a paramount prerequisite for success in complex spatial-temporal tasks.

    \item \textbf{The Necessity of Cohesive Summarization:} To further validate our design, we bypassed the intermediate summarization step, feeding raw clip captions directly to the QA module. This resulted in a consistent performance decline, suggesting that summarization is critical for distilling scattered visual cues from redundant local descriptions into a coherent global narrative for effective state tracking. Detailed results are in the Appendix.
\end{enumerate}
\section{Conclusion}
In this paper, we introduce \textbf{\benchmark}, a rigorous benchmark for evaluating multi-state visual reasoning in LMMs. Our extensive evaluation of 16 leading models reveals a significant performance gap compared to human cognition. Through a novel Text-Oracle diagnostic probe, we provide some critical insights regarding this task. By releasing \textbf{\benchmark}~and these critical insights, we aim to catalyze the development of next-generation models with a more robust understanding of our dynamic world.

\bibliographystyle{splncs04}
\bibliography{main}



\newpage
\appendix
\section{\benchmark~Details}

\begin{figure}[ht]
  \centering
  \begin{subfigure}[t]{0.6\linewidth}
    \centering
    \includegraphics[width=\linewidth]{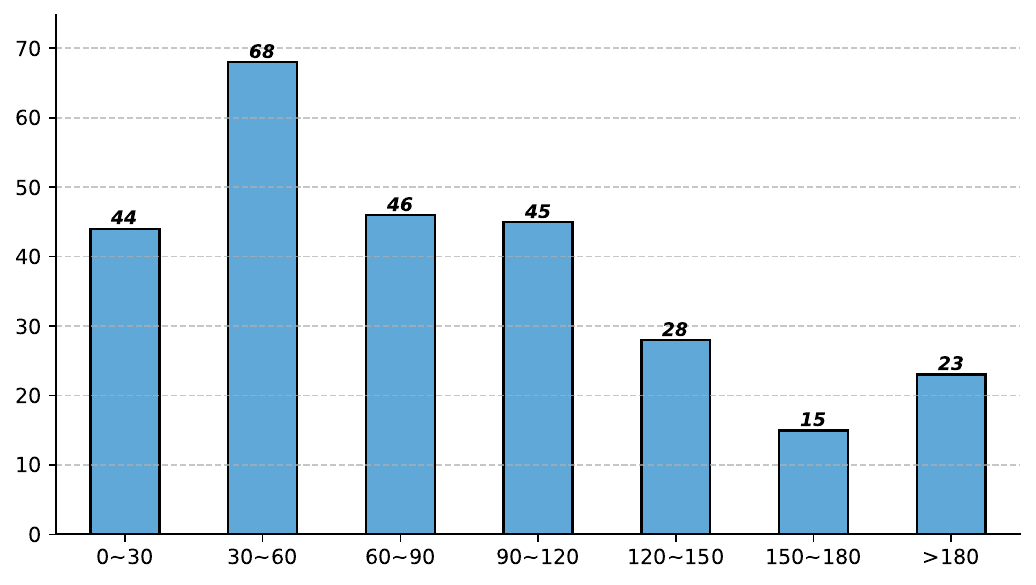}
    \caption{Distribution of scene areas. This chart shows the distribution of areas across all 270 scenes, highlighting the diversity of environmental scales included in the benchmark.}
    \label{fig:area}
  \end{subfigure}
  \hfill
  \begin{subfigure}[t]{0.35\linewidth}
    \centering
    \includegraphics[width=\linewidth]{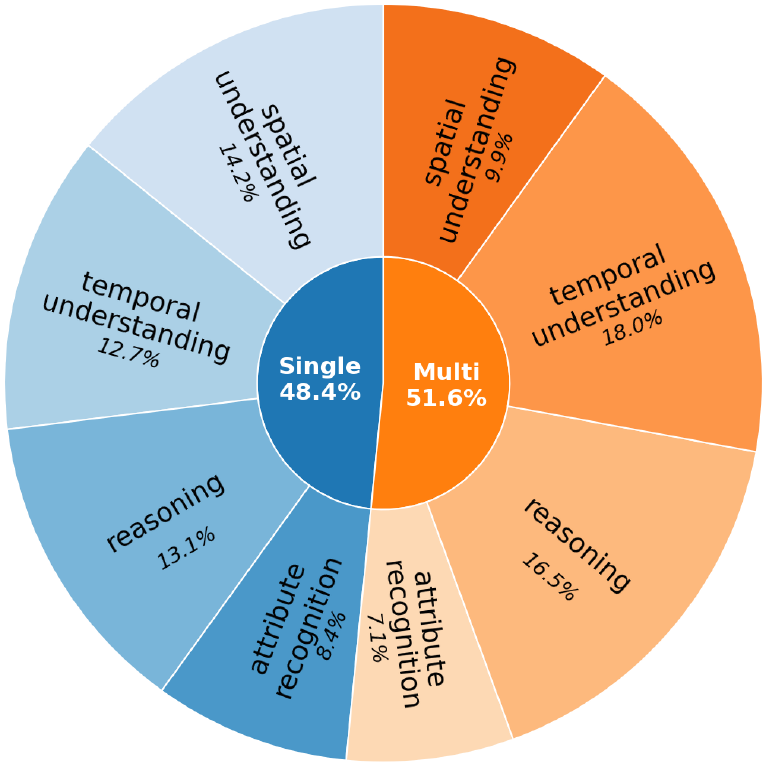}
    \caption{Question distribution by capability in \textbf{\benchmark}.}
    \label{fig:capabilities}
  \end{subfigure}
  \caption{Overview of \textbf{\benchmark} Statistics. (a) Distribution of scene areas across all 270 simulated scenes, demonstrating the environmental diversity of \textbf{\simsubset}. (b) Distribution of question capabilities across the entire benchmark, broken down by ``Intra-State'' and ``Inter-State'' question types.}
  \label{fig:combined}
\end{figure}

\subsection{\textbf{\simsubset}}
\textbf{\simsubset} is composed of 270 scenes derived from the ProcTHOR-10k dataset~\cite{ProcTHOR}. Each scene is documented with two egocentric, global observation videos that capture the environment in its initial and final states, bracketing a specific change. The resolution of videos is 1024$\times$768.

Beyond the video data, each scene is augmented with the following supplementary information, constituting a comprehensive resource for subsequent research:
\begin{enumerate}
    \item \textbf{Scene Layout Data}: Static information describing the environment's layout, which remains consistent across both the \textit{`before'} and \textit{`after'} states. This includes its total area and coordinates, along with the ID, area, coordinates, category, and adjacency information (room IDs) for each constituent room.
    \item \textbf{Depth Data}: Per-frame depth maps from the observation sequence to facilitate spatial awareness.
    \item \textbf{Instance Segmentation Data}: Per-frame instance segmentation maps from the observation sequence.
    \item \textbf{Object Data}: Comprehensive information for all objects across both the ``before" and ``after" states, encompassing each object's ID, category, position, orientation, parent receptacle, and attributes.
    \item \textbf{Operation Log}: A record of the specific operations applied to designated objects to enact the environmental change. It includes object ID, operation type and operation parameters.
\end{enumerate}

\subsubsection{Meta Data Collection}
\paragraph{\textbf{Trajectory Generation}}
Upon loading a scene from ProcTHOR-10k~\cite{ProcTHOR}, we first query the backend for all reachable points for the agent. The agent is then initialized at a random starting position chosen from this set. From this initial point, we generate a trajectory designed for the agent to efficiently explore and observe the environment.

While generating a plausible and efficient exploration trajectory for an embodied agent is a non-trivial challenge. A naive random walk is inefficient and offers no guarantee of comprehensive coverage, while an exhaustive traversal of all reachable points is computationally expensive and unnatural, involving redundant movements through open spaces. The ideal trajectory must balance observational completeness with path efficiency, mimicking how a human might systematically survey a new environment.

Inspired by human patrol strategies, our trajectory generation algorithm is based on a rectangular decomposition of the navigable space. The core principle is that large, open areas with many reachable points are often empty and do not require exhaustive traversal; it is more efficient to observe them by patrolling their perimeters. The algorithm proceeds as follows: first, we identify a set of maximal rectangles that cover as much of the navigable area as possible. These rectangles are constrained such that their area exceeds a predefined threshold and their entire perimeter consists of reachable points.

To enhance path efficiency and safety, the initial rectangles are contracted inward. A trajectory that closely follows the outer boundaries of the reachable coordinates is problematic for two main reasons. First, it results in unnecessarily long paths, reducing exploration efficiency. Second, it introduces a significant risk of collision with nearby objects or walls, which could disrupt the agent's movement. This contraction process may cause some larger rectangles to degenerate into simpler geometric primitives, such as line segments or single points. Finally, the shortest path connecting these contracted geometric shapes is computed using the A* search algorithm. This links all the primitives into a single, continuous tour, which forms the agent's complete exploration trajectory.

\paragraph{\textbf{Environment Changing}}
A distinguishing feature of our benchmark is its focus on inter-state scene understanding, making the generation of environmental changes a cornerstone of our methodology. To ensure the integrity and realism of our tasks, all modifications strictly adhere to the physical and semantic constraints inherent to the AI2-THOR~\cite{AI2-THOR} simulator. This rigorous process guarantees that every altered scene remains physically plausible and logically coherent. We introduce a diverse set of controlled modifications to simulate realistic alterations. These changes encompass: 
\begin{itemize}
    \item \textbf{Object Extrinsic Changes:} Altering the presence or pose of objects, including moving an object to a new location, adding a new object to the scene, or removing an existing one. When moving an object, we first identifies all receptacles within the scene where the object can be placed, along with the specific coordinates on those receptacles that can support the object. A position is then randomly selected from all the possible placement coordinates, and the object is moved to that location.
    \item \textbf{Object Intrinsic State Changes:} Modifying the internal state of an object through simulator-supported interactions:
    \begin{itemize}
        \item \textbf{Open/Close Object}: Applies to objects with the \texttt{openable} property(e.g. book, laptop). This operation allows for adjusting the openness of an object. 
        \item \textbf{Slice Object}: Applicable to objects possessing the \texttt{sliceable} property(e.g. apple, potato, tomato), provided they have not been sliced previously.
        \item \textbf{Break Object}: Targets objects marked as \texttt{breakable}(e.g. egg, bottle, bowl) that are currently intact.
        \item \textbf{Toggle Object On/Off}: Designed for \texttt{toggleable} objects(e.g., lights, fans), enabling the alteration of their binary toggled state.
        \item \textbf{Clean/Dirty Object}: Utilized for objects with the \texttt{dirtyable} property(e.g. bed, cloth, cup), allowing them to be cleaned if dirty or made dirty if clean.
        \item \textbf{Cook Object}: Applicable to \texttt{cookable} objects(e.g. egg, break, potato) that are in an uncooked state.
        \item \textbf{Fill/Empty Liquid}: Pertains to objects that \texttt{canFillWithLiquid}(e.g. bottle, bowl, cup). This includes filling an empty object with specified liquids (e.g., water, coffee, wine) or emptying an object that is currently filled.
        \item \textbf{Use Up Object}: Designed for objects that \texttt{canBeUsedUp}(e.g. toilet paper, soap bottle) and have not yet reached their ``used up" state.
    \end{itemize}
\end{itemize}

\paragraph{\textbf{Object-level Attributions Generation}}
The native object information provided by the simulator backend lacks essential visual attributes such as color and shape, which are critical for effective object referring and grounding. Furthermore, objects within the same category often correspond to different assets, each possessing unique visual properties. To address this limitation, we developed a pipeline to generate detailed attributes for each unique object asset.

\begin{figure}[ht]
\noindent
\begin{minipage}{\linewidth}
    \fcolorbox{gray!50!black}{white}{%
        \begin{minipage}{\dimexpr\linewidth-2\fboxsep-2\fboxrule\relax}
            \colorbox{gray!10}{%
                \parbox{\dimexpr\linewidth-10pt\relax}{%
                    \bfseries\itshape\scriptsize
                    DAM Prompt
                }%
            }%
            \vskip 2pt
            \hrule height 0.5pt
            \vskip 2pt
            \tiny
            The masked object is a(n) \{object\_category\}. Describe it in detail, focusing on its type, shape, color and physical properties.
            \vskip 2pt
        \end{minipage}%
    }
\end{minipage}
\caption{Prompt for Describe Anything Model (DAM).}
\label{fig:dam_prompt}
\end{figure}
\begin{figure*}[ht]
\noindent
\begin{minipage}{\linewidth}
    \fcolorbox{gray!50!black}{white}{%
        \begin{minipage}{\dimexpr\linewidth-2\fboxsep-2\fboxrule\relax}
            \colorbox{gray!10}{%
                \parbox{\dimexpr\linewidth-10pt\relax}{%
                    \bfseries\itshape\scriptsize
                    Consolidate Descriptions Prompt
                }%
            }%
            \vskip 2pt
            \hrule height 0.5pt
            \vskip 2pt
            
            \tiny
            \vskip 2pt
            You are given multiple descriptions of the same object. Please consolidate these descriptions into a single, comprehensive, and coherent description that:
            
            \vskip 4pt
            \textbf{1.} Integrates all consistent information from the multiple descriptions
            \par\vskip 2pt
            \textbf{2.} Resolves any contradictions by choosing the most frequently mentioned or most detailed information
            \par\vskip 2pt
            \textbf{3.} Fills in missing details by combining information from different descriptions
            \par\vskip 2pt
            \textbf{4.} Produces a more complete and reliable description than any individual description
            \par\vskip 4pt
            
            Here are the descriptions to consolidate: \\
            \{descriptions\_text\}\\
            
            \vskip 4pt
            Please provide a single consolidated description that is comprehensive, coherent, and incorporates the best information from all descriptions. Focus on physical properties, appearance and distinguishing features.
            
            \vskip 4pt
            \noindent\textbf{Consolidated description:}
            
        \end{minipage}%
    }
\end{minipage}
\caption{Prompt for consolidating multiple object descriptions.}
\label{fig:consolidate_prompt}
\end{figure*}

First, we isolate all visual instances of a specific asset by leveraging the recorded first-person RGB frames and their corresponding object segmentation masks. From the resulting image patches for each asset, we select the five highest-quality views—prioritizing those that are largest and closest to the center of the frame—to ensure a clear representation. These five images are then input into a Describe Anything Model~\cite{DAM} to generate distinct textual descriptions from various viewpoints with prompt shown in Fig.~\ref{fig:dam_prompt}. Subsequently, these multiple descriptions for a single asset are fed into an LLM~\cite{qwen3} to synthesize a unified and comprehensive final description for that asset with prompt shown in Fig.~\ref{fig:consolidate_prompt}.

\begin{figure*}[ht]
\noindent
\begin{minipage}{\linewidth}
    \fcolorbox{gray!50!black}{white}{%
        \begin{minipage}{\dimexpr\linewidth-2\fboxsep-2\fboxrule\relax}
            \colorbox{gray!10}{%
                \parbox{\dimexpr\linewidth-10pt\relax}{%
                    \bfseries\itshape\scriptsize
                    Comparative Analysis Prompt
                }%
            }%
            \vskip 2pt
            \hrule height 0.5pt
            \vskip 2pt
            
            \tiny
            \vskip 2pt
            
            You are given descriptions of \{num\_objects\} different \{object\_category\} objects. Please analyze and compare these objects to identify their distinguishing characteristics.
            
            \vskip 4pt
            \{objects\_text\}
            
            \vskip 4pt
            For each object, provide a JSON response that highlights what makes it unique compared to the others in the same category. Focus on distinguishing features like size, color and shape.
            
            \vskip 4pt
            For \texttt{shape} and \texttt{color}, the value should be a short phrase that fits grammatically in the following sentences:
            \begin{itemize}
                \item An object which is [shape]. (e.g., ``rectangular'', ``round'')
                \item An object which is [color]. (e.g., ``red'', ``brown'', ``green'', ``blue'')
            \end{itemize}
            
            \vskip 4pt
            Please respond with a JSON object where each key is the object ID and the value contains the object's distinctive attributes:
            
            \vskip 4pt
            \noindent
            \begin{minipage}{\linewidth}
                \tiny
                \texttt{\{}
                
                \hspace{2em}\texttt{"\{first\_asset\_id\}" : \{}
                
                \hspace{4em}\texttt{"shape" : "<1 word, adjective describing shape>",}
                
                \hspace{4em}\texttt{"color" : "<1 word, color description>",}
                
                \hspace{4em}\texttt{"other\_features" : "<2-4 words describing other notable features>",}
                
                \hspace{4em}\texttt{"description" : "<1-2 sentences describing this specific \{object\_category\} and what makes it different from the others>"}
                
                \hspace{2em}\texttt{\},}
                
                \hspace{2em}\texttt{"\{second\_asset\_id\}" : \{}
                
                \hspace{4em}\texttt{"shape" : "<1 word, adjective describing shape>",}
                
                \hspace{4em}\texttt{"color" : "<1 word, color description>",}
                
                \hspace{4em}\texttt{"other\_features" : "<2-4 words describing other notable features>",}
                
                \hspace{4em}\texttt{"description" : "<1-2 sentences describing this specific \{object\_category\} and what makes it different from the others>"}
                
                \hspace{2em}\texttt{\}}
                
                \texttt{\}}
            \end{minipage}
            
            \vskip 4pt
            \textbf{Instructions:}
            \begin{itemize}
                \item Focus on what makes each object unique within this category
                \item Ensure the values for \texttt{shape} and \texttt{color} are concise and fit the grammatical examples provided
                \item If any attribute cannot be determined, set its value to \texttt{null}
                \item Ensure descriptions emphasize distinguishing characteristics
                \item Output only valid JSON
            \end{itemize}
        \end{minipage}
    }
\end{minipage}
\caption{Prompt for comparative analysis of multiple objects.}
\label{fig:comparative_prompt}
\end{figure*}

Recognizing that assets within the same category share common functionalities but differ in appearance, we perform a final comparative analysis. The descriptions for several different assets belonging to the same category are provided to an LLM~\cite{qwen3} with prompt shown in Fig.~\ref{fig:comparative_prompt}. The model is prompted to compare them, identify the consistent object function, and extract the discriminative attributes (e.g., color, shape, and special features) that make each asset unique. Through this process, we successfully generated detailed descriptions and attribute profiles for \textbf{964} unique object assets.

\subsubsection{Question-Answer Pairs Generation}
\paragraph{\textbf{Robust Entity Grounding Framework}}
In natural language, entities are typically referenced by their general categories, such as ``the trash can" or ``the bedroom". However, real-world scenes often feature multiple instances of the same category, rendering such generic references ambiguous. Resolving this requires descriptive language precise enough to localize specific entities—for instance, distinguishing ``the red trash can" from a blue one, or specifying ``the larger bedroom". The challenge of generating unambiguous referring expressions is critical for evaluating fine-grained understanding in agents.

We tackles this challenge by implementing a multi-faceted, hierarchical grounding mechanism. This mechanism leverages a rich set of contextual cues to dynamically adjust the granularity of descriptions, prioritizing minimal yet sufficient information. By balancing conciseness with clarity, our approach prevents over-specification while ensuring unique identification, thereby generating highly specific references:

\begin{itemize}
    \item \textbf{Hierarchical Uniqueness Assessment:} We systematically evaluate the uniqueness of objects and rooms based on their intrinsic properties (e.g., \texttt{type}, \texttt{shape}, \texttt{size} and \texttt{color}) at both global (scene-wide) and local (type-specific) levels. This ensures that each entity is identified using the most concise unambiguous descriptor available.
    \item \textbf{Relational and Positional Grounding:} Beyond intrinsic properties, our system incorporates relational contexts, identifying entities through containment relationships (e.g., ``apple in fridge") and relative spatial ordering (e.g., ``the lowest object painting"). This capability is vital for distinguishing between visually similar entities.
\end{itemize}

A novel aspect of our approach is the ability to track uniquely identifiable entities across varying scene states. By comparing object sets and their properties between distinct states, we can pinpoint objects that have undergone changes or are uniquely present/absent. This facilitates the generation of questions related to dynamic scene evolution and object manipulation.

\paragraph{\textbf{Multi-Dimensional QA Design and Template-based Generation}}

Leveraging the rich scene metadata and our robust entity grounding, we employ an automated, template-based approach to construct the QA pairs. Our generation framework is structured along two primary dimensions to ensure comprehensive evaluation:

\begin{itemize}
    \item \textbf{Contextual Dimension (State Requirement):} Questions are first categorized based on the required context. \textbf{Intra-state} questions are designed to be answerable by observing only one of the two videos (either \textit{before} or \textit{after}), while \textbf{inter-state} questions necessitate comparing and reasoning across both to identify the change.
    
    \item \textbf{Capability Dimension (Cognitive Skills):} The templates are designed to probe a spectrum of core capabilities, including \textbf{spatial reasoning} (e.g., location, relations), \textbf{temporal understanding} (e.g., navigation ordering), \textbf{attribute identification} (e.g., color, state), and \textbf{comparative analysis} (e.g., counting, logical deduction).
\end{itemize}
By combining these dimensions, we generated templates for over \textbf{50 distinct subtasks}, spanning various granularities from object-level and room-level to scene-level queries. A key design principle is that individual questions are often composite, requiring a blend of these skills. For example, a question might require both spatial localization and attribute recognition. This multifaceted design introduces a unique challenge, compelling models to develop a holistic understanding of the environment while precisely localizing critical information across space and time.

\begin{table*}[ht]
\centering
\renewcommand{\arraystretch}{1.3} 
\tiny
\caption{Intra-State Question Templates for tasks in \textbf{\benchmark}}
\label{tab:intra_state_templates}

\resizebox{\textwidth}{!}{%
\begin{tabular}{|l|p{4.5cm}|p{4.2cm}|} 
\hline
\textbf{Category} & \textbf{Subtask} & \textbf{Template} \\
\hline
\multirow{8}{*}{scene\_info} &
  scene\_area &
  What is the area of this scene? \\
\cline{2-3}
 &
  room\_area &
  What is the area of \textcolor{red}{\{room\_name\}}? \\
\cline{2-3}
 &
  room\_type &
  What type of \textcolor{red}{\{room\_name\}} is? \\
\cline{2-3}
 &
  room\_function &
  What is \textcolor{red}{\{room\_name\}} used for? \\
\cline{2-3}
 &
  room\_shape &
  What is the shape of \textcolor{red}{\{room\_name\}}? \\
\cline{2-3}
 &
  scene\_room\_types &
  What types of rooms are in this scene? \\
\cline{2-3}
 &
  room\_type\_count &
  How many \textcolor{red}{\{room\_type\}} are there in this scene? \\
\cline{2-3}
 &
  room\_connectivity\_pair &
  Which room pairs are directly connected? \\
\hline
\multirow{12}{*}{object\_info} &
  object\_distance\_estimation &
  What is the distance between \textcolor{red}{\{object1\}} and \textcolor{red}{\{object2\}} at \textcolor{red}{\{state\}}? \\
\cline{2-3}
 &
  object\_distance\_shortest\_comparison &
  Which object is closer to \textcolor{red}{\{object\}} at \textcolor{red}{\{state\}}? \\
\cline{2-3}
 &
  object\_distance\_longest\_comparison &
  Which object is farthest from \textcolor{red}{\{object\}} at \textcolor{red}{\{state\}}? \\
\cline{2-3}
 &
  object\_shape &
  What is the shape of \textcolor{red}{\{object\_id\}} at \textcolor{red}{\{state\}}? \\
\cline{2-3}
 &
  object\_color &
  What color is \textcolor{red}{\{object\_id\}} at \textcolor{red}{\{state\}}? \\
\cline{2-3}
 &
  object\_height &
  What is the height of \textcolor{red}{\{object\_id\}} at \textcolor{red}{\{state\}}? \\
\cline{2-3}
 &
  object\_length &
  What is the length of \textcolor{red}{\{object\_id\}} at \textcolor{red}{\{state\}}? \\
\cline{2-3}
 &
  object\_room\_location &
  In which room is \textcolor{red}{\{object\_id\}} located at \textcolor{red}{\{state\}}? \\
\cline{2-3}
 &
  object\_type\_num\_in\_room &
  How many \textcolor{red}{\{object\_type\}} are there in \textcolor{red}{\{room\_name\}} at \textcolor{red}{\{state\}}? \\
\cline{2-3}
 &
  object\_type\_num\_in\_scene &
  How many \textcolor{red}{\{object\_type\}} are there in the entire scene at \textcolor{red}{\{state\}}? \\
\cline{2-3}
 &
  object\_receptacle\_location &
  What is \textcolor{red}{\{object\_id\}} placed on at \textcolor{red}{\{state\}}? \\
\cline{2-3}
 &
  receptacle\_contents &
  What objects are in \textcolor{red}{\{receptacle\_id\}} at \textcolor{red}{\{state\}}? \\
\hline
\multirow{14}{*}{agent\_explore} &
  room\_visit\_order &
  Which rooms were visited after \textcolor{red}{\{room\}} at \textcolor{red}{\{state\}}? \\
\cline{2-3}
 &
  object\_observation\_longest &
  Which object was observed for the longest time at \textcolor{red}{\{state\}}? \\
\cline{2-3}
 &
  object\_observation\_shortest &
  Which object was observed for the shortest time at \textcolor{red}{\{state\}}? \\
\cline{2-3}
 &
  room\_longest\_time &
  In which room did the observer stay for the longest time at \textcolor{red}{\{state\}}? \\
\cline{2-3}
 &
  room\_shortest\_time &
  In which room did the observer stay for the shortest time at \textcolor{red}{\{state\}}? \\
\cline{2-3}
 &
  object\_first\_appearance\_earliest &
  Among the following objects, which one was first observed earliest at \textcolor{red}{\{state\}}? \\
\cline{2-3}
 &
  object\_first\_appearance\_latest &
  Among the following objects, which one was first observed latest at \textcolor{red}{\{state\}}? \\
\cline{2-3}
 &
  object\_last\_appearance\_earliest &
  Among the following objects, which one was last observed earliest at \textcolor{red}{\{state\}}? \\
\cline{2-3}
 &
  object\_last\_appearance\_latest &
  Among the following objects, which one was last observed latest at \textcolor{red}{\{state\}}? \\
\cline{2-3}
 &
  room\_visit\_count &
  How many times did the observer visit \textcolor{red}{\{room\}} at \textcolor{red}{\{state\}}? \\
\cline{2-3}
 &
  nth\_visit\_room\_area &
  What is the area of the room \textcolor{red}{\{visit\_order\}} visited at \textcolor{red}{\{state\}}? \\
\cline{2-3}
 &
  nth\_visit\_room\_type &
  What is the type of the room \textcolor{red}{\{visit\_order\}} visited at \textcolor{red}{\{state\}}? \\
\cline{2-3}
 &
  nth\_visit\_room\_shape &
  What is the shape of the room \textcolor{red}{\{visit\_order\}} visited at \textcolor{red}{\{state\}}? \\
\cline{2-3}
 &
  nth\_visit\_room\_objects &
  Which objects are observed in the room \textcolor{red}{\{visit\_order\}} visited at \textcolor{red}{\{state\}}? \\
\hline
\end{tabular}
}
\end{table*}
\begin{table*}[ht]
\centering
\renewcommand{\arraystretch}{1.3}
\tiny
\caption{Inter-State Question Templates for \textbf{object\_change} tasks in \textbf{\benchmark}}
\label{tab:inter_state_templates_object_change}

\resizebox{\textwidth}{!}{%
\begin{tabular}{|l|p{4.5cm}|p{4.2cm}|} 
\hline
\textbf{Category} & \textbf{Subtask} & \textbf{Template} \\
\hline
\multirow{22}{*}{object\_change} &
  movement\_distance &
  How far was the \textcolor{red}{\{object\_name\}} moved from \textcolor{red}{\{state1\}} to \textcolor{red}{\{state2\}}? \\
\cline{2-3}
 &
  old\_receptacle &
  Where was \textcolor{red}{\{object\}} at \textcolor{red}{\{state1\}} located at \textcolor{red}{\{state2\}}? \\
\cline{2-3}
 &
  new\_receptacle &
  Where is \textcolor{red}{\{object\}} at \textcolor{red}{\{state2\}} located at \textcolor{red}{\{state1\}}? \\
\cline{2-3}
 &
  movement\_between\_receptacles &
  Where did \textcolor{red}{\{object\}} move from \textcolor{red}{\{state1\}} to \textcolor{red}{\{state2\}}? \\
\cline{2-3}
 &
  new\_visible\_objects &
  Which objects became visible between \textcolor{red}{\{state1\}} and \textcolor{red}{\{state2\}}? \\
\cline{2-3}
 &
  lost\_visible\_objects &
  Among \textcolor{red}{\{state1\}} and \textcolor{red}{\{state2\}}, which objects are no longer visible? \\
\cline{2-3}
 &
  objects\_room\_movement &
  Which of the following objects changed rooms between \textcolor{red}{\{state1\}} and \textcolor{red}{\{state2\}}? \\
\cline{2-3}
 &
  longest\_observed\_object\_type &
  Among \textcolor{red}{\{state1\}} and \textcolor{red}{\{state2\}}, which object has the longest total observation time? \\
\cline{2-3}
 &
  shortest\_observed\_object\_type &
  Among \textcolor{red}{\{state1\}} and \textcolor{red}{\{state2\}}, which object has the shortest total observation time? \\
\cline{2-3}
 &
  largest\_diff\_observation\_time &
  Among \textcolor{red}{\{state1\}} and \textcolor{red}{\{state2\}}, which object has the largest difference in observation time between the two states? \\
\cline{2-3}
 &
  smallest\_diff\_observation\_time &
  Among \textcolor{red}{\{state1\}} and \textcolor{red}{\{state2\}}, which object has the smallest difference in observation time between the two states? \\
\cline{2-3}
 &
  attribute\_changed\_objects &
  Which of the following objects had any attribute changes between \textcolor{red}{\{state1\}} and \textcolor{red}{\{state2\}} (excluding positional movements)? \\
\cline{2-3}
 &
  room\_attribute\_changed\_objects &
  In \textcolor{red}{\{room\}}, which of the following objects had attribute changes between \textcolor{red}{\{state1\}} and \textcolor{red}{\{state2\}} (excluding positional movements)? \\
\cline{2-3}
 &
  room\_position\_changes &
  In \textcolor{red}{\{room\}}, which objects changed position between \textcolor{red}{\{state1\}} and \textcolor{red}{\{state2\}}? \\
\cline{2-3}
 &
  openness\_state\_comparison &
  What was the openness state of \textcolor{red}{\{object\_name\}} in \textcolor{red}{\{state1\}} and \textcolor{red}{\{state2\}}? \\
\cline{2-3}
 &
  toggle\_state\_comparison &
  What was the power state of \textcolor{red}{\{object\_name\}} in \textcolor{red}{\{state1\}} and \textcolor{red}{\{state2\}}? \\
\cline{2-3}
 &
  cleanliness\_state\_comparison &
  What was the cleanliness state of \textcolor{red}{\{object\_name\}} in \textcolor{red}{\{state1\}} and \textcolor{red}{\{state2\}}? \\
\cline{2-3}
 &
  cooking\_state\_comparison &
  What was the cooking state of \textcolor{red}{\{object\_name\}} in \textcolor{red}{\{state1\}} and \textcolor{red}{\{state2\}}? \\
\cline{2-3}
 &
  liquid\_state\_comparison &
  What was the liquid filling state of \textcolor{red}{\{object\_name\}} in \textcolor{red}{\{state1\}} and \textcolor{red}{\{state2\}}? \\
\cline{2-3}
 &
  breaking\_state\_comparison &
  What was the condition of \textcolor{red}{\{object\_name\}} in \textcolor{red}{\{state1\}} and \textcolor{red}{\{state2\}}? \\
\cline{2-3}
 &
  slicing\_state\_change &
  Was \textcolor{red}{\{object\_name\}} sliced between \textcolor{red}{\{state1\}} and \textcolor{red}{\{state2\}}? \\
\cline{2-3}
 &
  usage\_state\_change &
  Was \textcolor{red}{\{object\_name\}} used up between \textcolor{red}{\{state1\}} and \textcolor{red}{\{state2\}}? \\
\hline
\end{tabular}%
}
\end{table*}
\begin{table*}[ht]
\centering
\renewcommand{\arraystretch}{1.3}
\tiny 
\caption{Inter-State Question Templates for \textbf{agent\_explore} tasks in \textbf{\benchmark}}
\label{tab:inter_state_templates_agent_explore}

\resizebox{\textwidth}{!}{%
\begin{tabular}{|l|p{4.5cm}|p{4.2cm}|} 
\hline
\textbf{Category} & \textbf{Subtask} & \textbf{Template} \\
\hline
\multirow{14}{*}{agent\_explore} &
  longest\_stay\_room &
  In which room did the observer stay for the longest time among \textcolor{red}{\{state1\}} and \textcolor{red}{\{state2\}}? \\
\cline{2-3}
 &
  shortest\_stay\_room &
  In which room did the observer stay for the shortest time among \textcolor{red}{\{state1\}} and \textcolor{red}{\{state2\}}? \\
\cline{2-3}
 &
  cross\_state\_longest\_room\_comparison &
  Among the following options, which state-room combination had the longest observation time? \\
\cline{2-3}
 &
  cross\_state\_shortest\_room\_comparison &
  Among the following options, which state-room combination had the shortest observation time? \\
\cline{2-3}
 &
  room\_visit\_count\_comparison &
  Which room was visited more times in \textcolor{red}{\{state1\}} compared to \textcolor{red}{\{state2\}}? \\
\cline{2-3}
 &
  total\_room\_visits &
  How many times did the observer visit \textcolor{red}{\{room\}} in total across \textcolor{red}{\{state1\}} and \textcolor{red}{\{state2\}}? \\
\cline{2-3}
 &
  object\_observation\_comparison &
  Which object was observed for a longer total time across \textcolor{red}{\{state1\}} and \textcolor{red}{\{state2\}}? \\
\cline{2-3}
 &
  rooms\_visited\_both\_states &
  Which rooms were visited in both \textcolor{red}{\{state1\}} and \textcolor{red}{\{state2\}}? \\
\cline{2-3}
 &
  rooms\_visited\_only\_state1 &
  Which rooms were visited only in \textcolor{red}{\{state1\}} but not in \textcolor{red}{\{state2\}}? \\
\cline{2-3}
 &
  rooms\_visited\_only\_state2 &
  Which rooms were visited only in \textcolor{red}{\{state2\}} but not in \textcolor{red}{\{state1\}}? \\
\cline{2-3}
 &
  nth\_mth\_visit\_room\_type &
  Do the \textcolor{red}{\{visit1\_order\}} visited room in the \textcolor{red}{\{state1\}} video and the \textcolor{red}{\{visit2\_order\}} visited room in the \textcolor{red}{\{state2\}} video have the same type? If not, what are their respective types? \\
\cline{2-3}
 &
  nth\_mth\_visit\_room\_shape &
  Do the \textcolor{red}{\{visit1\_order\}} visited room in the \textcolor{red}{\{state1\}} video and the \textcolor{red}{\{visit2\_order\}} visited room in the \textcolor{red}{\{state2\}} video have the same shape? If not, what are their respective shapes? \\
\cline{2-3}
 &
  nth\_mth\_visit\_room\_size &
  Which room has a larger area: the \textcolor{red}{\{visit1\_order\}} visited room in the \textcolor{red}{\{state1\}} video or the \textcolor{red}{\{visit2\_order\}} visited room in the \textcolor{red}{\{state2\}} video? And approximately what percentage larger is it? \\
\cline{2-3}
 &
  nth\_mth\_visit\_room\_same &
  Are the \textcolor{red}{\{visit1\_order\}} visited room in the \textcolor{red}{\{state1\}} video and the \textcolor{red}{\{visit2\_order\}} visited room in the \textcolor{red}{\{state2\}} video the same room? \\
\hline
\end{tabular}%
}
\end{table*}

Question generation templates are shown in Tab.~\ref{tab:intra_state_templates},  Tab.~\ref{tab:inter_state_templates_object_change}, and 
Tab.~\ref{tab:inter_state_templates_agent_explore}.

\paragraph{\textbf{LLM Filtering and Refinement}}

To ensure our benchmark consists of high-quality questions that are natural, visually-grounded, and appropriately challenging, we refine the initial template-based QA pairs using a two-stage, LMM-driven filtering process:
\begin{itemize}
    \item \textbf{Stage 1: Text-based Filtering} This stage processes QA pairs without any visual context.
    \begin{itemize}
        \item \textbf{Vision-free Answerability Check:} We ask the LLM~\cite{qwen3} to effectively determine if a question can be reliably answered using only the text of the question, options, and ground-truth answer with prompt shown in Fig.~\ref{fig:vision-free_filter_prompt}. Questions that do not require visual information are discarded. For example, a question like ``What is the function of the bed?" with the answer ``For sleeping" would be removed, as it tests general knowledge rather than visual perception of the scene.
        
        \item \textbf{Ambiguity Removal:} The LMM~\cite{qwen3} evaluates whether the questions or options contain ambiguities or potential confusions with prompt shown in Fig.~\ref{fig:ambiguity_removal_prompt} and such ambiguous cases are removed. For instance, a question asking the model to distinguish between two objects with nearly identical colors (e.g., options like ``the cyan cup" versus ``the light blue cup") would be flagged and eliminated, as the distinction is subjective and unreliable.
        
        \item \textbf{Linguistic Enhancement:} The LLM rewrites the template-based questions and options into more natural, fluent, and human-like expressions with prompt shown in Fig.~\ref{fig:linguistic_enhancement_prompt}.
    \end{itemize}
    
    \item \textbf{Stage 2: Multimodal Filtering} We conduct preliminary answer testing using a lightweight LMM with full vision-language context. For each question, we run multiple inference trials. If the model consistently answers correctly without additional prompting, we deem the question too easy and remove it for lacking sufficient challenge.
\end{itemize}

\begin{figure*}[h]
\noindent
\begin{minipage}{\linewidth}
    \fcolorbox{gray!50!black}{white}{%
        \begin{minipage}{\dimexpr\linewidth-2\fboxsep-2\fboxrule\relax}
            \colorbox{gray!10}{%
                \parbox{\dimexpr\linewidth-10pt\relax}{%
                    \bfseries\itshape\scriptsize
                    Vision-free Answerability Check Prompt
                }%
            }%
            \vskip 2pt
            \hrule height 0.5pt
            \vskip 2pt
            \tiny
            Your task is to determine whether a question requires visual information to answer.\\
            
            Question: \{question\}\\
            Options: \{options\}\\
            Answer: \{answer\}\\
            
            If the question can be answered correctly without visual information, respond with ``Filter out". Otherwise, respond with ``Keep". Your response must be either ``Filter out" or ``Keep", with no additional text or explanation.
            \vskip 2pt
        \end{minipage}%
    }
\end{minipage}
\caption{Vision-free Answerability Check Prompt.}
\label{fig:vision-free_filter_prompt}
\end{figure*}
\begin{figure}[t]
\noindent
\begin{minipage}{\linewidth}
    \fcolorbox{gray!50!black}{white}{%
        \begin{minipage}{\dimexpr\linewidth-2\fboxsep-2\fboxrule\relax}
            \colorbox{gray!10}{%
                \parbox{\dimexpr\linewidth-10pt\relax}{%
                    \bfseries\itshape\scriptsize
                    Ambiguity Removal Prompt
                }%
            }%
            \vskip 2pt
            \hrule height 0.5pt
            \vskip 2pt
            \tiny
            \vskip 2pt
            
            You are a QA quality control assistant. Your task is to filter out QAs where the options are ambiguous. Your evaluation should solely focus on identifying if the options are unclear or ambiguous based on the provided text and options. Do not consider visual aspects, question wording, or correctness of the answer at this stage. Evaluate the following question, options, and correct answer.\\
            
            Question: \{question\}\\
            Options: \{options\}\\
            Answer: \{answer\}\\
            
            If you identify any issues (e.g., ambiguous options), respond with ``Filter out". Otherwise, respond with ``Keep". Your response must be either ``Filter out" or ``Keep", with no additional text or explanation.
        \end{minipage}
    }
\end{minipage}
\caption{Ambiguity Removal Prompt.}
\label{fig:ambiguity_removal_prompt}
\end{figure}
\begin{figure}[t]
\noindent
\begin{minipage}{\linewidth}
    \fcolorbox{gray!50!black}{white}{%
        \begin{minipage}{\dimexpr\linewidth-2\fboxsep-2\fboxrule\relax}

            \colorbox{gray!10}{%
                \parbox{\dimexpr\linewidth-10pt\relax}{%
                    \bfseries\itshape\scriptsize
                    Linguistic Enhancement Prompt
                }%
            }%
            \vskip 2pt
            \hrule height 0.5pt
            \vskip 2pt

            \tiny
            \vskip 2pt
            
            You are a helpful assistant. Here you are provided with a question and its options generated with templates, which may contain unnatural phrasing. Your task is to rewrite the given question and options to make them sound more like natural language without changing their semantic meaning. Do not change the order of the options. Pay special attention to rewriting expressions like ``state\_0" and ``state\_1" to be more natural and closer to real language. Provide the rewritten question and options in JSON format with keys ``rewritten\_question" and ``rewritten\_options".\\
            
            Question: \{question\}\\
            Options: \{options\}\\
            
            Example Output:
            
            \vskip 4pt
            \noindent
            \tiny
            \texttt{\{}
            
            \hspace{2em}\texttt{"rewritten\_question": "What is the capital of France?",}
            
            \hspace{2em}\texttt{"rewritten\_options": ["Paris", "London", "Berlin"]}
            
            \texttt{\}}
            \normalsize
            
            \vskip 4pt
        \end{minipage}
    }
\end{minipage}
\caption{Linguistic Enhancement Prompt.}
\label{fig:linguistic_enhancement_prompt}
\end{figure}

Through this pipeline, we ultimately constructed a high-quality, high-discriminability inter-state visual question answering benchmark.
\paragraph{\textbf{Human Reviewing, Testing and Cleansing}}
We recruited 12 reviewers with at least a bachelor's degree to re-examine and screen the data, and to evaluate human performance on \textbf{\simsubset}~and ultimately yielded \textbf{2,932 high-quality QA pairs}.

\subsection{\textbf{\realsubset}}
While \textbf{\simsubset}~provides a controlled environment for fine-grained diagnostic analysis, the construction of \textbf{\realsubset} serves three critical purposes:
\begin{itemize}
    \item to closely approximate the visual complexity, lighting variations, and physical noise inherent in real-world scenarios;
    \item to enhance the overall credibility of the benchmark by validating the findings on physical data devoid of simulation artifacts; 
    \item to rigorously evaluate whether a Sim-to-Real Gap exists for modern LMMs. 
\end{itemize}

By juxtaposing performance on synthetic and real data, we aim to diagnose potential distribution shifts and assess the true generalization capacity of these models in dynamic physical environments.

\paragraph{\textbf{Data Collection}}
To ensure the data reflects the intricacies of physical reality, we recruited human actors to record videos using handheld mobile devices in previously unseen indoor environments—specifically selected to avoid overlap with any existing public datasets. This deliberate design ensures that the scenes in \textbf{\realsubset} are novel and independent, minimizing the risk of data leakage and preventing models from leveraging prior exposure to identical or highly similar scenes during training.

Actors were instructed to film in landscape orientation, following distinct exploration trajectories to observe the scene in both the ``before'' and ``after'' states. To simulate natural human observation while maintaining visual stability, they moved at a uniform horizontal speed. We strictly enforced protocols to avoid excessive camera shaking, drastic vertical movements, or deliberately focusing on the modified objects to prevent visual bias. The dataset encompasses \textbf{29 indoor scenes} with diverse styles, captured under varying lighting conditions and with different hardware devices to maximize environmental complexity.

State transitions were physically executed by human annotators, resulting in transformations ranging from 4 to over 20 distinct modifications per scene. These manipulations strictly adhered to physical plausibility, ensuring that changes in object states are both realistic and human-executable.

\paragraph{\textbf{Question-Answer Pairs Construction}}
To ensure the diversity of linguistic expression and mitigate potential cognitive bias, the QA pairs were designed collaboratively by 12 experienced annotators. This multi-annotator strategy ensures that the questions reflect a broad spectrum of reasoning patterns and natural language variations, rather than being confined to the specific phrasing style of a few individuals.

The construction of the QA pairs follows the taxonomy and task categories established in the \textbf{\simsubset}~generation pipeline. However, due to the absence of backend metadata in real-world scenes, we employ auxiliary methods to guarantee answer accuracy. For questions involving spatial metrics, we utilize architectural floor plans and manual measurements to obtain ground-truth annotations. To maintain consistency with the synthetic subset, we rigorously apply the \textbf{Robust Entity Grounding Framework} to resolve referential ambiguity. By fusing fine-grained visual attributes with precise spatial context, we ensure that each question unambiguously refers to a specific entity (e.g., distinguishing between identical objects based on their unique relational properties).

The initial annotation pool consists of approximately 680 questions. To curate a high-quality, ``shortcut-resistant'' benchmark, we employ a model-in-the-loop filtering strategy. Candidate questions that can be answered correctly by all evaluated models are discarded as trivial. Through this rigorous selection process, we retain a final set of \textbf{552 challenging QA pairs}.

\subsection{Cases}
\begin{figure*}[t]
    \centering
    \includegraphics[width=1\linewidth]{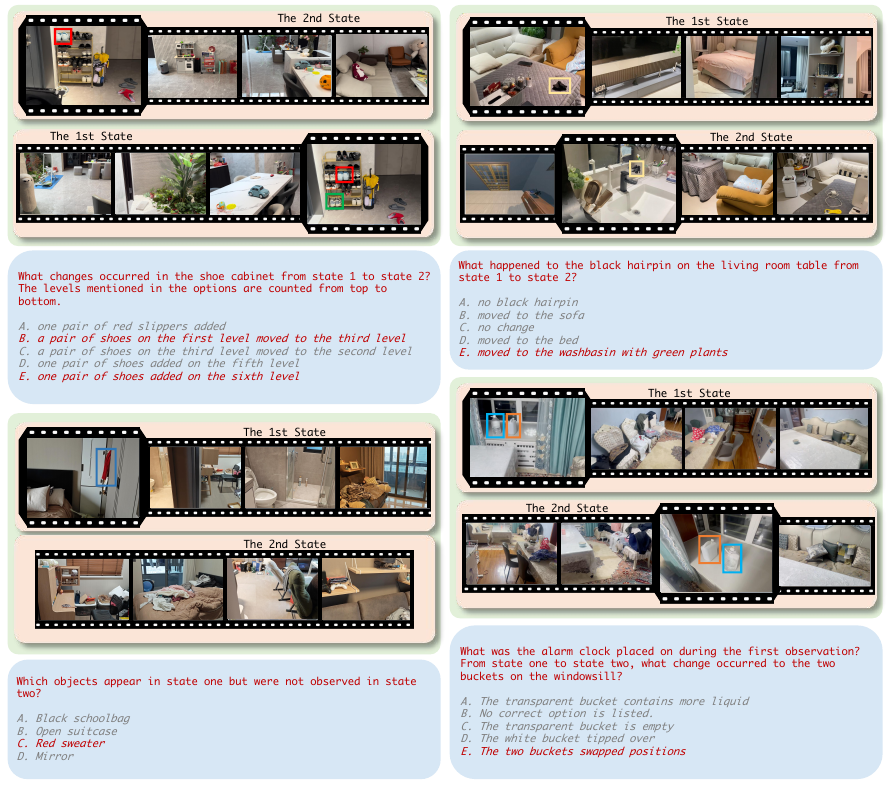}
    \caption{\textbf{\realsubset} Examples}
    \label{fig:real_examples}
\end{figure*}

\begin{figure*}[t]
    \centering
    \includegraphics[width=1\linewidth]{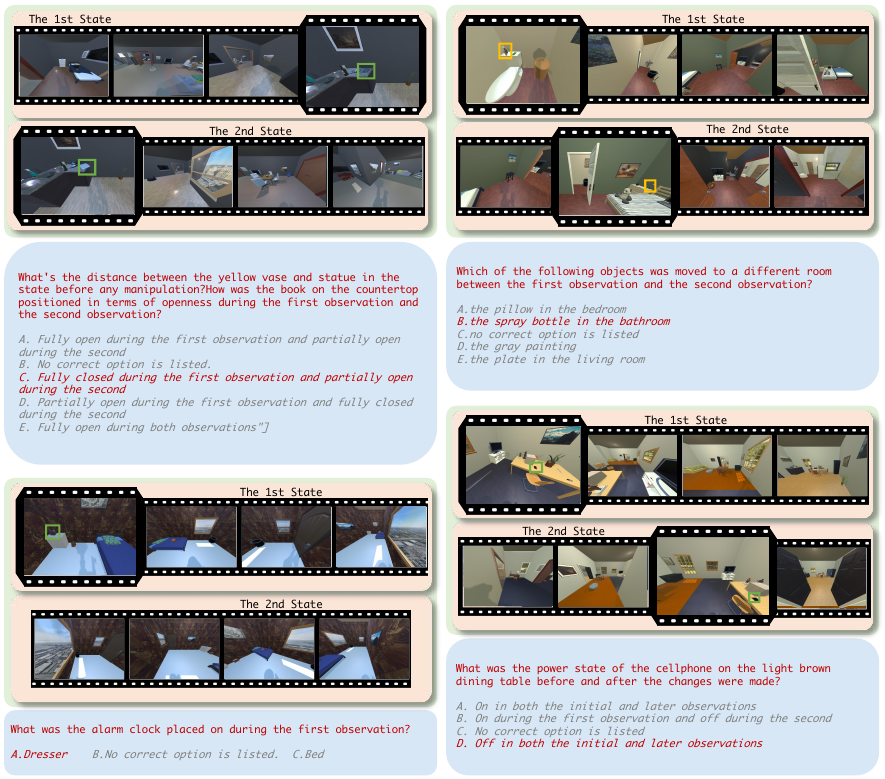}
    \caption{\textbf{\simsubset} Examples}
    \label{fig:sim_examples}
\end{figure*}

Some QA examples are shown in Fig.~\ref{fig:real_examples} and Fig.~\ref{fig:sim_examples}.

\section{Evaluation Details}

\subsection{Test Setting}

\paragraph{Evaluation Strategies.}
To ensure fair and rigorous comparisons across models with diverse video processing capabilities, we implement a standardized evaluation protocol. The protocol is detailed as follows:

\begin{itemize}
    \item \textbf{Input Frames.} For models accessed through an API, our methodology was to maximize the frame count within the constraints of their supported context windows. This guiding principle led to the final configuration of 200 frames per video for GPT-5, a frame rate of 2 fps for Gemini-2.5-Pro, and 1 fps for Qwen3-VL-235B-A22B-Instruct. For models executed locally, the number of input frames was tailored to their contextual capabilities. The default protocol involved uniformly sampling 200 frames from each of the \textit{`before'} and \textit{`after'} state videos, resulting in a total of 400 frames being input to the LMM. Notably, for the InternVL3.5 series~\cite{InternVL3.5}, InternVideo2.5~\cite{InternVideo2.5} and VideoLLaMA3~\cite{VideoLLaMA3}, their restricted context lengths necessitated a reduction to 60 sampled frames per video (120 frames total). Video-XL-2~\cite{Video-XL-2} demanded a bespoke adjustment due to its inherent lack of support for multi-video inputs. For this model, we concatenated the frames from both videos into a continuous stream, incorporating explicit textual markers to delineate the \textit{`before'} and \textit{`after'} content.

    \item \textbf{Input Resolution.} For closed-source models, the resolution at which video frames are processed is an internal parameter fixed by the service provider, and therefore cannot be adjusted by the user. For open-source models, the input resolution is standardized to 448$\times$448 for all models to maintain consistency. The only exception is VideoLLaMA3~\cite{VideoLLaMA3}, which is set to 336$\times$448 resolution to accommodate its fixed aspect ratio requirements.

    \item \textbf{Thinking-mode Parameters.} To standardize the reasoning process across models, we configured model-specific inference parameters where available. For the API-based models, we set the \texttt{reasoning\_effort} to \texttt{low} for GPT-5 and the \texttt{thinking\_budget} to \texttt{128} for Gemini-2.5-Pro. Qwen3-VL-235B-A22B-Instruct was called in a \texttt{non-thinking} mode. For the local models, we disabled explicit thinking modes for all models, with the sole exception of WeThink-Qwen2.5VL-7B~\cite{WeThink}, which was evaluated using its native, built-in thinking mode.

\end{itemize}

\paragraph{Evaluating Prompt.}
\begin{figure}[t]
\noindent
\begin{minipage}{\linewidth}
    \fcolorbox{gray!50!black}{white}{%
        \begin{minipage}{\dimexpr\linewidth-2\fboxsep-2\fboxrule\relax}
            \colorbox{gray!10}{%
                \parbox{\dimexpr\linewidth-10pt\relax}{%
                    \bfseries\itshape\scriptsize
                    LLM Evaluation Prompt
                }%
            }%
            \vskip 2pt
            \hrule height 0.5pt
            \vskip 2pt
            \tiny
            Your task is to answer questions based on the provided videos and options.\\
            
            Question: \{question text\}\\
            Options: \{candidate options\}\\
            
            Return ONLY the letter(s) corresponding to the correct option(s) (e.g., A, B, C, or A,C for multiple correct answers)\\
            
            Questions with only one correct option: There is only one correct option for this question.\\
            Questions with multiple correct options: There are more than one correct option for this question.
            \vskip 2pt
        \end{minipage}%
    }
\end{minipage}
\caption{Evaluation Prompt for Vision-Blind LLMs~\cite{qwen3}.}
\label{fig:llm_evaluation_prompt}
\end{figure}
\begin{figure}[t]
\noindent
\begin{minipage}{\linewidth}
    \fcolorbox{gray!50!black}{white}{%
        \begin{minipage}{\dimexpr\linewidth-2\fboxsep-2\fboxrule\relax}
            \colorbox{gray!10}{%
                \parbox{\dimexpr\linewidth-10pt\relax}{%
                    \bfseries\itshape\scriptsize
                    LMM Evaluation Prompt
                }%
            }%
            \vskip 2pt
            \hrule height 0.5pt
            \vskip 2pt
            \tiny
            Your task is to answer questions based on the provided videos and options.\\
            
            This is the first video: \{video of the `before' state\}\\
            This is the second video: \{video of the `after' state\}\\
            
            Question: \{question text\}\\
            Options: \{candidate options\}\\
            
            Return ONLY the letter(s) corresponding to the correct option(s) (e.g., A, B, C, or A,C for multiple correct answers)\\
            
            Questions with only one correct option: There is only one correct option for this question.\\
            Questions with multiple correct options: There are more than one correct option for this question.
            \vskip 2pt
        \end{minipage}%
    }
\end{minipage}
\caption{Evaluation Prompt for LMMs.}
\label{fig:lmm_evaluation_prompt}
\end{figure}
When evaluating the LLMs~\cite{qwen3}, we only input the question, options, and necessary prompts, requiring the model to answer based solely on the text input, as shown in~\ref{fig:llm_evaluation_prompt}. When evaluating the LMMs, in addition to the LLM inputs, we also provided videos showing the before and after states of the scene corresponding to the question, requiring the model to answer based on the video content, as shown in~\ref{fig:lmm_evaluation_prompt}.

The prompt provided to each model is dynamically adjusted to specify whether the question is single-choice or multiple-choice. This decision was based on preliminary experiments, which revealed that without such guidance, a subset of models exhibited a tendency to hallucinate multiple answers even for questions with only a single correct option. This behavior would lead to undeserved penalties and an inaccurate evaluation of their true capabilities. Therefore, to mitigate this issue and ensure a fair assessment, we explicitly instruct the model on the expected number of answers based on the ground truth.

\subsection{Zero-shot Evaluation Results}
\paragraph{Evaluation Metrics} Since the benchmark includes both single-choice and multiple-choice questions, we score the model's answers based on accuracy: if the model's answer is a subset of the ground truth, it receives a score proportional to the fraction of the ground truth it covers; if the model's selection includes any incorrect option, it receives zero points for that question.

\section{Text-Oracle Evaluation Details}

\subsection{Prompts}

\begin{figure}[ht]
\noindent
\begin{minipage}{\linewidth}
    \fcolorbox{gray!50!black}{white}{%
        \begin{minipage}{\dimexpr\linewidth-2\fboxsep-2\fboxrule\relax}
            \colorbox{gray!10}{%
                \parbox{\dimexpr\linewidth-10pt\relax}{%
                    \bfseries\itshape\scriptsize
                    Clip Caption Generation Prompt
                }%
            }%
            \vskip 2pt
            \hrule height 0.5pt
            \vskip 2pt
            \tiny
            \vskip 2pt
            
            This is a video clip segment from a video.\\
            Total video duration: \{total length of the video\}\\
            This video clip segment duration time: \{timestamps\}\\
            
            Please describe the video content in detail.\\
            Focus on the environment spatial layout, objects, including their spatial layout, detailed information, and the flow of the video.\\
            Please describe in clear and concise language.
        \end{minipage}
    }
\end{minipage}
\caption{Clip Caption Generation Prompt.}
\label{fig:clip_caption}
\end{figure}
\begin{figure}[h]
\noindent
\begin{minipage}{\linewidth}
    \fcolorbox{gray!50!black}{white}{%
        \begin{minipage}{\dimexpr\linewidth-2\fboxsep-2\fboxrule\relax}
            \colorbox{gray!10}{%
                \parbox{\dimexpr\linewidth-10pt\relax}{%
                    \bfseries\itshape\scriptsize
                    Captions Summarization Prompt
                }%
            }%
            \vskip 2pt
            \hrule height 0.5pt
            \vskip 2pt
            \tiny
            \vskip 2pt
            
            Please synthesize the following descriptions of different segments of a video into a single, coherent, and detailed overall caption for the entire video.\\
            
            Segment Descriptions:\\
            \{captions for all the clips that make up this complete video\}\\
            
            Focus on the environment spatial layout, objects, including their spatial layout, detailed information, and the flow of the video.\\
            Please describe in clear and concise language.
        \end{minipage}
    }
\end{minipage}
\caption{Captions Summarization Prompt.}
\label{fig:caption_summarization}
\end{figure}
\begin{figure}[h]
\noindent
\begin{minipage}{\linewidth}
    \fcolorbox{gray!50!black}{white}{%
        \begin{minipage}{\dimexpr\linewidth-2\fboxsep-2\fboxrule\relax}
            \colorbox{gray!10}{%
                \parbox{\dimexpr\linewidth-10pt\relax}{%
                    \bfseries\itshape\scriptsize
                    Text-Oracle Answer Prompt
                }%
            }%
            \vskip 2pt
            \hrule height 0.5pt
            \vskip 2pt
            \tiny
            \vskip 2pt
            
            Based on the following context information, answer the question:\\
            Context: \{context\}\\
            Question: \{question\}\\
            Options: \{options\_text\}\\
            
            Return ONLY the option letter(s) (e.g., "A", "B", "C"). For multiple correct options, concatenate the letters (e.g., "ABC"). AVOID ANY OTHER OUTPUT CONTENT.
        \end{minipage}
    }
\end{minipage}
\caption{Text-Oracle Answer Prompt.}
\label{fig:Text-Oracle_answer}
\end{figure}

Prompts used in Text-Oracle are shown in Fig.~\ref{fig:clip_caption}, Fig.~\ref{fig:caption_summarization} and Fig.~\ref{fig:Text-Oracle_answer}

\subsection{Ablation Study}

We performed an ablation study on the Text-Oracle method to evaluate the effect of varying clip lengths on different models. For InternVL3.5-4B and InternVL3.5-8B~\cite{InternVL3.5}, we assessed clip lengths of 50, 100, and 200, uniformly sampling 50 frames from each clip. For the Qwen3-VL series, we evaluated clip lengths of 100 and 200 for the Qwen3-VL-4B model and a length of 200 for the Qwen3-VL-8B model, with each clip being uniformly sampled to 100 frames.

In testing the Text-Oracle method, we also conducted an ablation study by removing the caption summarization module with InternVL3.5-4B and InternVL3.5-8B~\cite{InternVL3.5}. In this configuration, the large model was tasked with generating answers directly from the raw captions of each clip. Clip length is set as 50. The results is shown in Tab.~\ref{tab:Text-Oracle_without_summary}

\begin{table*}[t]
\centering
\renewcommand{\arraystretch}{1.08}
\footnotesize
\caption{Ablation study of different clip length in Text-Oracle.}
\label{tab:Text-Oracle}
\resizebox{\textwidth}{!}{%
\begin{tabular}{l|c|cccccccccc}
\toprule
   &
   &
  \multicolumn{5}{c|}{\textbf{Intra-state}} &
  \multicolumn{5}{c}{\textbf{Inter-state}} \\
\cmidrule(lr){3-7} \cmidrule(lr){8-12}
  \multirow{-2}{*}{\textbf{Model}} &
  \multirow{-2}{*}{\textbf{Overall}} &
  \textbf{Avg.} &
  \textbf{S. U.} &
  \textbf{T. U.} &
  \textbf{A. R.} &
  \textbf{Rs} &
  \textbf{Avg.} &
  \textbf{S. U.} &
  \textbf{T. U.} &
  \textbf{A. R.} &
  \textbf{Rs} \\
\midrule

\multicolumn{12}{l}{\textit{InternVL3.5-4B-Instruct}~\cite{InternVL3.5}} \\
\midrule
  \rowcolor[HTML]{F4B084} 
  InternVL3.5-4B-Instruct~\cite{InternVL3.5} & 35.47 & 32.63 & 32.99 & 33.39 & 34.17 & 30.52 & 38.46 & 39.78 & 34.62 & 45.51 & 38.78 \\
  \rowcolor[HTML]{F8CBAD} 
  InternVL3.5-4B-Instruct-clip50~\cite{InternVL3.5}  & 38.08 & 32.27 & 34.14 & 30.62 & 33.39 & 31.14 & 44.32 & 46.15 & 40.45 & 51.95 & 44.14 \\
  \rowcolor[HTML]{F8CBAD} 
  InternVL3.5-4B-Instruct-clip100~\cite{InternVL3.5} & 38.37 & 32.31 & 33.41 & 30.90 & 33.87 & 31.50 & 44.86 & 47.58 & 41.07 & 51.54 & 44.47 \\
  \rowcolor[HTML]{F8CBAD} 
  InternVL3.5-4B-Instruct-clip200~\cite{InternVL3.5} & 38.67 & 31.65 & 33.58 & 30.52 & 30.55 & 31.38 & 45.78 & 47.21 & 42.11 & 54.25 & 45.26 \\
\midrule

\multicolumn{12}{l}{\textit{InternVL3.5-8B-Instruct}~\cite{InternVL3.5}} \\
\midrule
  \rowcolor[HTML]{9BC2E6} 
  InternVL3.5-8B-Instruct~\cite{InternVL3.5} & 31.04 & 31.01 & 31.35 & 33.20 & 31.86 & 27.97 & 32.81 & 32.47 & 33.93 & 32.01 & 32.14 \\
  \rowcolor[HTML]{BDD7EE} 
  InternVL3.5-8B-Instruct-clip50~\cite{InternVL3.5}  & 37.62 & 32.25 & 33.68 & 30.65 & 33.18 & 31.66 & 44.53 & 48.99 & 40.33 & 51.65 & 43.34 \\
  \rowcolor[HTML]{BDD7EE} 
  InternVL3.5-8B-Instruct-clip100~\cite{InternVL3.5} & 38.53 & 32.49 & 33.77 & 31.26 & 32.56 & 32.27 & 45.77 & 50.11 & 42.59 & 50.02 & 44.77 \\
  \rowcolor[HTML]{BDD7EE} 
  InternVL3.5-8B-Instruct-clip200~\cite{InternVL3.5} & 36.90 & 31.80 & 33.34 & 29.84 & 33.83 & 30.75 & 43.09 & 46.05 & 39.33 & 49.95 & 42.42 \\
\midrule

\multicolumn{12}{l}{\textit{Qwen3-VL-4B-Instruct}~\cite{qwen3-vl}} \\
\midrule
  \rowcolor[HTML]{A9D08E} 
  Qwen3-VL-4B-Instruct~\cite{qwen3-vl}         & 38.05 & 34.84 & 34.42 & 36.95 & 39.35 & 30.36 & 41.77 & 40.65 & 40.38 & 53.24 & 39.00 \\
  \rowcolor[HTML]{C6E0B4} 
  Qwen3-VL-4B-Instruct-clip100~\cite{qwen3-vl} & 38.69 & 34.93 & 36.52 & 33.08 & 36.47 & 34.01 & 42.15 & 45.25 & 38.94 & 48.07 & 41.23 \\
  \rowcolor[HTML]{C6E0B4} 
  Qwen3-VL-4B-Instruct-clip200~\cite{qwen3-vl} & 38.48 & 34.29 & 35.83 & 32.90 & 35.33 & 33.32 & 42.67 & 45.12 & 39.32 & 49.67 & 41.82 \\
\midrule

\multicolumn{12}{l}{\textit{Qwen3-VL-8B-Instruct}~\cite{qwen3-vl}} \\
\midrule
  \rowcolor[HTML]{FFD966} 
  Qwen3-VL-8B-Instruct~\cite{qwen3-vl}         & 35.87 & 35.82 & 33.29 & 38.56 & 40.82 & 32.71 & 37.46 & 40.06 & 37.77 & 36.33 & 36.05 \\
  \rowcolor[HTML]{FFE699} 
  Qwen3-VL-8B-Instruct-clip200~\cite{qwen3-vl} & 38.54 & 34.06 & 35.18 & 32.30 & 36.43 & 33.05 & 42.95 & 45.55 & 39.68 & 50.63 & 41.61 \\
\bottomrule
\end{tabular}
}
\end{table*}
\begin{table*}[t]
\centering
\renewcommand{\arraystretch}{1.08}
\footnotesize
\caption{Text-Oracle without caption summarization with clip\_len = 50.}
\label{tab:Text-Oracle_without_summary}
\resizebox{\textwidth}{!}{%
\begin{tabular}{l|c|cccccccccc}
\toprule
   &
   &
  \multicolumn{5}{c|}{\textbf{Intra-state}} &
  \multicolumn{5}{c}{\textbf{Inter-state}} \\
\cmidrule(lr){3-7} \cmidrule(lr){8-12}
  \multirow{-2}{*}{\textbf{Model}} &
  \multirow{-2}{*}{\textbf{Overall}} &
  \textbf{Avg.} &
  \textbf{S. U.} &
  \textbf{T. U.} &
  \textbf{A. R.} &
  \textbf{Rs} &
  \textbf{Avg.} &
  \textbf{S. U.} &
  \textbf{T. U.} &
  \textbf{A. R.} &
  \textbf{Rs} \\
\midrule
\multicolumn{12}{l}{\textit{InternVL3.5-4B-Instruct}~\cite{InternVL3.5}} \\
\midrule
    \rowcolor[HTML]{F4B084} 
    InternVL3.5-4B-Instruct~\cite{InternVL3.5}            & 35.47 & 32.63 & 32.99 & 33.39 & 34.17 & 30.52 & 38.46 & 39.78 & 34.62 & 45.51 & 38.78 \\
    \rowcolor[HTML]{F8CBAD} 
    InternVL3.5-4B-Instruct~\cite{InternVL3.5}+Text-Oracle       & 38.67 & 31.65 & 33.58 & 30.52 & 30.55 & 31.38 & 45.78 & 47.21 & 42.11 & 54.25 & 45.26 \\
    \rowcolor[HTML]{FCE4D6} 
    InternVL3.5-4B-Instruct~\cite{InternVL3.5}+Qwen3-0.6B~\cite{qwen3} & 25.79 & 24.68 & 26.01 & 23.66 & 28.09 & 22.06 & 28.31 & 30.11 & 27.62 & 27.32 & 28.42 \\
    \rowcolor[HTML]{FCE4D6} 
    InternVL3.5-4B-Instruct~\cite{InternVL3.5}+Qwen3-4B~\cite{qwen3}   & 36.35 & 31.62 & 33.54 & 29.34 & 32.09 & 31.47 & 40.41 & 43.02 & 38.08 & 45.03 & 39.39 \\
    \rowcolor[HTML]{FCE4D6} 
    InternVL3.5-4B-Instruct~\cite{InternVL3.5}+Qwen3-8B~\cite{qwen3}   & 35.92 & 31.20 & 32.52 & 29.82 & 31.10 & 31.16 & 40.43 & 43.83 & 37.58 & 44.5  & 39.73 \\
    \rowcolor[HTML]{FCE4D6} 
    InternVL3.5-4B-Instruct~\cite{InternVL3.5}+Qwen3-14B~\cite{qwen3}  & 34.28 & 29.89 & 30.89 & 27.14 & 30.65 & 31.00 & 36.55 & 37.97 & 34.27 & 42.11 & 35.76 \\
    \rowcolor[HTML]{FCE4D6} 
    InternVL3.5-4B-Instruct~\cite{InternVL3.5}+Qwen3-32B~\cite{qwen3}  & 37.15 & 31.35 & 33.33 & 28.96 & 32.24 & 30.96 & 41.49 & 42.94 & 38.66 & 48.25 & 40.78 \\
\midrule
\multicolumn{12}{l}{\textit{InternVL3.5-8B-Instruct}~\cite{InternVL3.5}} \\
\midrule
    \rowcolor[HTML]{9BC2E6} 
    InternVL3.5-8B-Instruct~\cite{InternVL3.5}            & 31.04 & 31.01 & 31.35 & 33.20  & 31.86 & 27.97 & 32.81 & 32.47 & 33.93 & 32.01 & 32.14 \\
    \rowcolor[HTML]{BDD7EE} 
    InternVL3.5-8B-Instruct~\cite{InternVL3.5}+Text-Oracle       & 38.53 & 32.49 & 33.77 & 31.26 & 32.56 & 32.27 & 45.77 & 50.11 & 42.59 & 50.02 & 44.77 \\
    \rowcolor[HTML]{DDEBF7} 
    InternVL3.5-8B-Instruct~\cite{InternVL3.5}+Qwen3-0.6B~\cite{qwen3} & 25.95 & 26.50  & 26.91 & 26.55 & 29.34 & 24.21 & 27.26 & 29.86 & 26.50  & 26.84 & 26.69 \\
    \rowcolor[HTML]{DDEBF7} 
    InternVL3.5-8B-Instruct~\cite{InternVL3.5}+Qwen3-4B~\cite{qwen3}   & 36.26 & 31.94 & 34.33 & 29.64 & 31.17 & 32.08 & 40.29 & 42.83 & 37.56 & 45.83 & 39.33 \\
    \rowcolor[HTML]{DDEBF7} 
    InternVL3.5-8B-Instruct~\cite{InternVL3.5}+Qwen3-8B~\cite{qwen3}   & 36.12 & 31.51 & 33.88 & 29.11 & 31.24 & 31.44 & 40.74 & 43.85 & 37.55 & 46.03 & 40.06 \\
    \rowcolor[HTML]{DDEBF7} 
    InternVL3.5-8B-Instruct~\cite{InternVL3.5}+Qwen3-14B~\cite{qwen3}  & 33.84 & 29.18 & 32.63 & 25.04 & 29.62 & 29.18 & 36.48 & 37.76 & 34.15 & 41.96 & 35.89 \\
    \rowcolor[HTML]{DDEBF7} 
    InternVL3.5-8B-Instruct~\cite{InternVL3.5}+Qwen3-32B~\cite{qwen3}  & 36.34 & 31.36 & 33.90 & 27.92 & 32.46 & 31.24 & 40.40 & 42.19 & 37.02 & 48.04 & 39.68 \\
\bottomrule
\end{tabular}%
}
\end{table*}

\end{document}